\documentclass[preprint,10pt,3p,sort]{elsarticle}
\usepackage{graphics} % for pdf, bitmapped graphics files
\usepackage{epsfig} % for postscript graphics files
\usepackage{times} % assumes new font selection scheme installed
\usepackage{amsmath} % assumes amsmath package installed
\usepackage{amssymb}  % assumes amsmath package installed
\usepackage{epstopdf}
\usepackage{subfigure,color,balance}
\usepackage{caption}
\usepackage{verbatim}
\usepackage{cases}
\allowdisplaybreaks[4]
\usepackage{enumerate}
\usepackage{booktabs}
\usepackage{tabularx} 
\usepackage{balance}
\usepackage{mathbbol}
\usepackage{algorithm}
\usepackage{algorithmicx}
\usepackage{algpseudocode}
\usepackage{mathrsfs}
\usepackage{threeparttable}
\usepackage[draft]{todonotes}
\usepackage{makecell}

\usepackage{enumitem}

\newcommand{\tabincell}[2]{\begin{tabular}{@{}#1@{}}#2\end{tabular}} 
\newcommand*{\circled}[1]{\lower.7ex\hbox{\tikz\draw (0pt, 0pt)%
		circle (.5em) node {\makebox[1em][c]{\small #1}};}}
\usepackage[colorlinks=true,      % false: boxed links; true: colored links
linkcolor=black,      % color of internal links
citecolor=black,      % color of links to bibliography
filecolor=black,      % color of file links
urlcolor=blue]{hyperref}

\newdefinition{remark}{Remark}

\def\0{{\bf 0}}
\def\1{{\bf 1}}

 \biboptions{compress}
\begin{document}
	
	\begin{frontmatter}
		
		%\title{Data-driven predictive control for mixed traffic flow: Experimental validation in dissipating traffic waves}
		%\title{Cooperative data-driven predictive control for distributed wave mitigation in large-scale mixed traffic flow} 
		%\title{Distributed data-driven predictive control for cooperative wave mitigation \\ in large-scale mixed traffic flow}
		%\title{Smoothing large-scale mixed traffic flow: cooperative data-driven control and distributed implementation}
		\title{
        From Optimizable to Interactable: Mixed Digital Twin-Empowered \\Testing of Vehicle-Infrastructure Cooperation Systems
        } 
		%\tnotetext[mytitlenote]{Fully documented templates are available in the elsarticle package on \href{http://www.ctan.org/tex-archive/macros/latex/contrib/elsarticle}{CTAN}.}
		
		%% Group authors per affiliation:
		%\author{Chaoyi Chen, Qing Xu, Jiawei Wang, Jianqiang Wang and Keqiang Li}
		%%\fnref{myfootnote}
		%\address{School of Vehicle and Mobility, Tsinghua University, Beijing 100084, China}
		%%\fntext[myfootnote]{Since 1880.}
		
		%% or include affiliations in footnotes:
		\author[a]{Jianghong Dong}
		\ead{djh20@mails.tsinghua.edu.cn}
        \author[a]{Chunying Yang}
		\ead{ycyacademic@gmail.com}
		
        \author[a]{Mengchi Cai}
	  \ead{caimengchi@tsinghua.edu.cn}
        \author[a]{Chaoyi Chen}
		\ead{chency2023@tsinghua.edu.cn}

        \author[a]{Qing Xu\corref{cor1}}
		\ead{qingxu@tsinghua.edu.cn}
        \author[a]{Jianqiang Wang}
		\ead{wjqlws@tsinghua.edu.cn}
		\author[a]{Keqiang Li}
		\ead{likq@tsinghua.edu.cn}
  %       \author[b]{Kaidi Yang}
		% \ead{kaidi.yang@nus.edu.sg}

		\cortext[cor1]{Corresponding author: Qing Xu}
		
		%\fnref{fn1}
		%\fntext[fn1]{This is the first author footnote.}

		\address[a]{School of Vehicle and Mobility, Tsinghua University, 100084 Beijing, China}
		% \address[b]{Department of Civil and Environmental Engineering, National University of Singapore, 119077, Singapore}
		
\begin{abstract}
Sufficient testing under corner cases is critical for the long-term operation of vehicle-infrastructure cooperation systems (VICS). 
However, existing corner-case generation methods are primarily AI-driven, and VICS testing under corner cases is typically limited to simulation. 
In this paper, we introduce an L5 ``Interactable'' level to the VICS digital twin (VICS-DT) taxonomy, extending beyond the conventional L4 ``Optimizable'' level.
We further propose an L5-level VICS testing framework, IMPACT (Interactive Mixed-digital-twin Paradigm for Advanced Cooperative vehicle-infrastructure Testing). 
By enabling direct human interactions with VICS entities, IMPACT incorporates highly uncertain and unpredictable human behaviors into the testing loop, naturally generating high-quality corner cases that complement AI-based methods. 
Furthermore, the mixedDT-enabled ``Physical-Virtual Action Interaction'' facilitates safe VICS testing under corner cases, incorporating real-world environments and entities rather than purely in simulation. 
Finally, we implement IMPACT on the I-VIT (Interactive Vehicle-Infrastructure Testbed), and experiments demonstrate its effectiveness.
The experimental videos are available at our \href{https://dongjh20.github.io/IMPACT}{project website}.

\end{abstract}
		
		
	\end{frontmatter}
	
\section{Introduction}
\label{sec-introduction}

With breakthroughs in Vehicle-to-Everything (V2X) technologies driven by next-generation 5G communication,  ultra-reliable and low-latency communication (URLLC) among vehicles, roadside infrastructure, and cloud has become feasible~\cite{gao2025vehicle}. 
This further facilitates the development and practical deployment of vehicle-infrastructure cooperation systems (VICS)~\cite{yu2022review,zhang2025evaluation,liu2021towards} and even vehicle-road-cloud integration systems~\cite{li2020Principles,chu2021cloud}. 
Typical applications include cooperative perception to broaden sensing ranges~\cite{huang2025vehicle}, cooperative planning to reduce travel time~\cite{han2026cooperative}, and cooperative control to lower energy consumption~\cite{gao2025cloud}. 
VICS has already demonstrated great potential for improving the safety, efficiency, and sustainability of road traffic systems.

%Similar to Connected and Autonomous Vehicles (CAVs), 
Naturally, VICS requires extensive validation and testing before practical deployment~\cite{feng2021intelligent,feng2023dense}. 
Given that real-world field testing is prohibitively resource-intensive~\cite{xu2021system}, existing VICS studies mainly rely on simulation-based testing, such as infrastructure-based warning system that enhances driving safety~\cite{zhang2025evaluation}, smart streetlights that autonomously switch on and off according to vehicle positions~\cite{zhao2022regional} and intelligent intersections that dynamically adjust signal phases according to traffic flow~\cite{han2026cooperative,li2024cooperative}. 
However, pure simulations often fail to accurately reproduce the unmodeled and unpredictable noises and disturbances inherent in the physical world, such as low-level vehicle dynamics, sensor measurement biases, and communication jitters. 

\begin{figure}[t]
	\vspace{1mm}
	\centering
	\subfigure[Classical DT]
	{\includegraphics[scale=0.4]{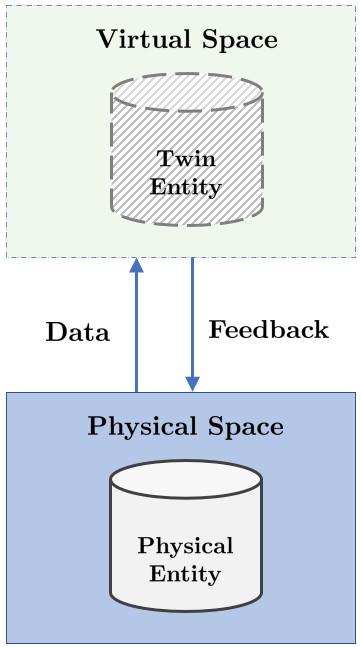}
		\label{fig-dt-architecture}}
	\hspace{9mm}
	\subfigure[MixedDT]
	{\includegraphics[scale=0.4]{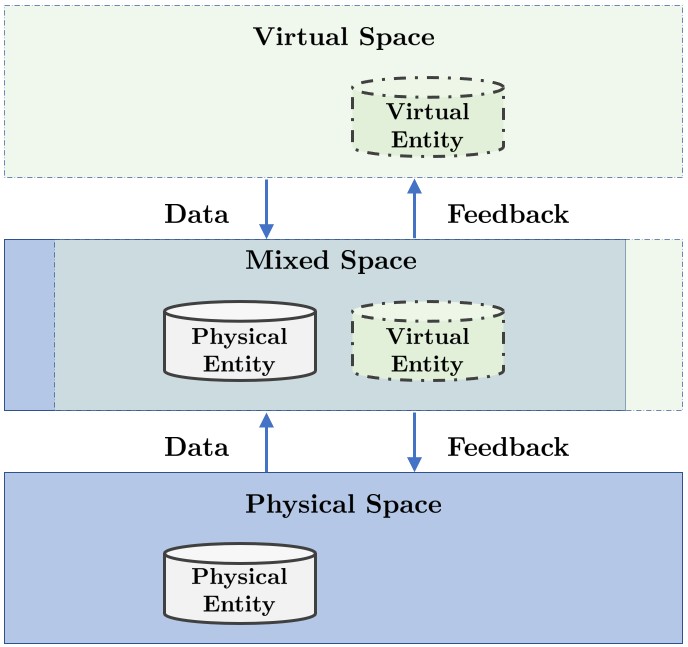}
		\label{fig-mdt-architecture}}
	\vspace{-1mm}
	\caption{Schematics for classical DT and mixedDT. (a) In classical DT, virtual entities are typically digital counterparts of the physical ones. (b) In mixedDT, virtual entities exist independently and the physical and virtual entities are integrated into the unified mixed space, where they can coexist and interact with each other.
	}
	\label{fig.architecture}
\end{figure}

To address these limitations, it is imperative to incorporate physical factors into virtual simulations. 
In this context, the Digital Twin (DT) framework~\cite{glaessgen2012digital}, which facilitates seamless connectivity and interaction between the physical and virtual entities, emerges as a highly promising tool. 
As illustrated in Fig.~\ref{fig-dt-architecture}, DT typically consists of two primary components: physical entities in the physical space and the virtual counterparts of the physical entities (twin entities) in the virtual space, which are linked by a bidirectional, closed-loop interaction. 
Through such bidirectional synchronization and interaction, DT promises
to facilitate VICS testing~\cite{gu2025digital}. Recently, several studies have begun to explore DT-based VICS testing in typical applications, including smart parking~\cite{zhang2022bsdp}, road infrastructure deployment optimization~\cite{demiyanushko2021virtual}, intelligent tunnel lighting~\cite{zhao2022regional}, and adaptive traffic signal control~\cite{dasgupta2024harnessing}.

\subsection{DT-based VICS Testing}

\begin{table}[t!]
    \centering
    \caption{Digital Twin Levels of VICS}
    \label{tab-VICS-DT-levels}
    
    \begin{tabularx}{\textwidth}{@{} l p{5.5cm} >{\raggedright\arraybackslash}X @{}}
        \toprule
        \textbf{Level} & \textbf{Capability} & \textbf{Requirement} \\
        \midrule
        \makecell[l]{L1 Visualizable \\ \cite{su2022individual,xu2023end}} & 
        Mirroring the physical via the virtual (Holographic Perception) & 
        Emerging perception data, requiring holographic data governance. \\
        \addlinespace
        
        \makecell[l]{L2 Diagnosable \\ \cite{chen2019evaluating,muppalla2017knowledge}} & 
        Diagnosing the physical via the virtual (Causal Inference) & 
        Spatio-temporally continuous detection data, requiring causal diagnosis of the road traffic system. \\
        \addlinespace
        
        \makecell[l]{L3 Predictable \\ \cite{zhu2022kst,yu2021trajectory}} & 
        Predicting the physical via the virtual (Decision Support) & 
        Numerous control strategies with high effect uncertainty, requiring decision support. \\
        \addlinespace
        
        \makecell[l]{L4 Optimizable \\ \cite{zhao2022regional,kamal2024digital}} & 
        Optimizing the physical via the virtual (Autonomic Control) & 
        Two-way data interaction, requiring autonomic closed-loop traffic control. \\
        % \midrule 
        \cmidrule(lr){1-3}

        % L5 整行加粗
        \textbf{L5 Interactable$^{\ast}$} & 
        \textbf{Actuating the physical via the virtual (Actionable Interaction)} & 
        \textbf{Bidirectional action-level interaction between physical and virtual entities, requiring direct human intervention in their control.
        } \\
        \bottomrule
        
        % 表格底部的注解
        \addlinespace[0.5ex]
        \multicolumn{3}{@{}l}{\footnotesize \textit{Note:} Levels L1--L4 are adapted from Sun et al. \cite{sun2023digital}. Level L5 ($^{\ast}$) is newly proposed in this study.} \\
    \end{tabularx}
\end{table}

Maturity taxonomy provides a structured way to understand existing research and anticipates future trends.
Unlike autonomous driving, which has the widely adopted SAE levels~\cite{sae2021taxonomy}, no broadly accepted taxonomy yet exists for VICS-DT. 
Therefore, as presented in Table~\ref{tab-VICS-DT-levels}, we adapt the taxonomy proposed by Sun et al.~\cite{sun2023digital}, which is originally developed for DT of road traffic systems and effectively captures the prevailing understanding and emerging trends in existing studies.
As VICS is a subset of road traffic systems, this taxonomy can be adapted to VICS.
Since the foundation of DT-based VICS testing inherently relies on the VICS-DT system itself, the maturity levels of VICS-DT can explicitly provide a clear understanding of how DT is applied in VICS testing, what role it plays, and what future directions it may take.
The following analyzes existing studies according to this taxonomy.

As illustrated in Table.~\ref{tab-VICS-DT-levels}, the original taxonomy only includes Levels L1-L4~\cite{sun2023digital}, and this paper proposes an extended Level 5 (L5) tailored to the practical requirements of VICS testing: 
\begin{itemize}
    \item L1 (Visualizable) focuses on mapping physical entities into the virtual space using observable data, which requires accurate and sufficient perception data. Typical topics include include vehicle trajectory reconstruction in roadside blind spots~\cite{su2022individual} and joint multi-vehicle detection and tracking~\cite{xu2023end}. L1 forms the foundation of VICS-DT construction.
    
\item L2 (Diagnosable) identifies problems and their causes in traffic systems based on data analysis, which requires model formulation and causal inference. 
Typical topics include diagnosing intersection efficiency caused by poor signal timing~\cite{chen2019evaluating}, and building knowledge graphs for traffic event diagnosis~\cite{muppalla2017knowledge}. 
L2 represents the initial application stage of VICS-DT.

\item L3 (Predictable) predicts future traffic states from existing data to support control decisions, which requires accurate prediction models. 
Typical topics include traffic flow forecasting~\cite{zhu2022kst} and high-risk event prediction on freeways~\cite{yu2021trajectory}. L3 transitions VICS-DT to real-time online operation.

\item L4 (Optimizable) optimizes physical control schemes through real-time, bidirectional data interaction between physical and virtual entities, which requires a closed-loop data interaction and real-time optimization strategies. Typical topics include dynamic tunnel lighting adjustment~\cite{zhao2022regional} and adaptive signal control at multiple intersections~\cite{kamal2024digital}. 
L4 is the current focus of VICS-DT research, generating optimization recommendations in the virtual space based on data uploaded by physical entities.

\end{itemize}
 
However, under the classic DT framework, virtual entities typically server as digital counterparts of physical ones. Consequently, the potential of virtual entities in DT-based VICS testing has not been fully explored and utilized in aforementioned studies. 
In particular, one major challenge in VICS testing is evaluation under corner cases, as such cases are most likely to cause system failures but are hard to capture in large-scale routine tests. 
Current corner case generation relies heavily on AI-driven methods~\cite{feng2023dense,feng2021intelligent}, which are limited by inherent flaws like hallucinations and fail to accurately simulate the highly unpredictable behaviors of human participants.
Moreover, since corner cases are typically dangerous and safety-critical scenarios, conventional VICS corner-case testing has predominantly relied on simulations~\cite{feng2023dense}, where real-world noise and disturbances cannot be accurately reproduced, resulting in incomplete and less reliable evaluation under corner cases.
To address these limitations, this paper aims to develop an VICS testing approach advancing beyond the L4 VICS-DT by leveraging the state-of-the-art Mixed Digital Twin (mixedDT) framework~\cite{dong2023mixed}.

\subsection{Interactable DT-based VICS testing}

As illustrated in Fig.~\ref{fig-mdt-architecture}, mixedDT is an extension of the classical DT framework, comprising the physical, virtual, and mixed spaces. 
In particular, virtual entities within the virtual space are not restricted to being digital counterparts of physical entities; instead, they can exist independently and directly interact with physical entities within the mixed space. 
Such interaction is not limited to data interaction (exchange), but extended to action-level interaction, meaning that the behavior of virtual entities can directly affect the state of physical entities, and vice versa.
Note that the physical space is a real-world testing environment, while the virtual space is highly configurable and reproducible, which is particularly applicable in safety-critical corner cases. 
The mixed space bridges the physical and virtual spaces to operate together, and thus facilitates more potential applications, enhancing the flexibility and scalability of DT's applications in VICS testing.

Based on mixedDT, and considering the limitations of existing corner case generation methods in VICS testing, this paper further proposes that VICS-DT should go beyond L4 (Optimizable) and introduce a higher level, namely L5 (Interactable); see Table~\ref{tab-VICS-DT-levels} for illustration. 
Specifically, beyond L4, where interaction between physical and virtual entities is mainly at the data level, the L5-level VICS-DT allows virtual entities to exist independently and engage in bidirectional action-level interaction with physical entities. 
Consequently, human operators can impose control intentions on both physical and virtual entities to alter their states, thereby directly interacting with them. 
We designate such a DT system as the Interactive Digital Twin (InteractiveDT).
Under this design, highly uncertain, unpredictable, and hard-to-model human behaviors can be directly incorporated into VICS testing, thereby naturally producing diverse high-quality corner cases. 
This effectively complements AI-driven corner case generation and enables more comprehensive testing of VICS before practical deployment.
Moreover, this enhances experimental flexibility by deploying hazardous entities virtually while keeping others physical, thereby enabling VICS corner-case testing that safely incorporates physical environments and entities.

%A related concept with mixedDT is Mixed Reality (MR), which is defined as a class of simulators combining both physical and virtual objects to create a hybrid of the physical and virtual spaces~\cite{rokhsaritalemi2020review}. MR evolves from multiple concepts including Augmented Reality and Augmented Virtuality. It aims to construct the “Reality-Virtuality Continuum” and consequently provides consumers a more immersive and interactive environment~\cite{flavian2019impact}. 
%Recently, MR has been applied to CAVs testing and evaluation to extend the existence of interactive objects in not only virtual testing environments~\cite{feng2023effect} but also physical testing environments~\cite{feng2020safety}. To address the practical testing demands of CAVs, mixedDT introduces the idea of MR into classical DT by establishing a mixed space where both physical and virtual entities could coexist and interact with each other. 

%将来可以加概念对比，和CPS中的虚实交互相比，和硬件在环的概念相比。。。as for Cyber Physical Systems (CPS)~\cite{lee2008cyber,rajkumar2010cyber}  This represents the successful realization of ``Physical-Virtual Action Interaction'', moving  beyond conventional data-level exchange, namely ``Physical-Virtual Data Interaction''.

\begin{figure}[t]
	%\vspace{-3mm}
	\centering
	\includegraphics[width=0.9\textwidth]{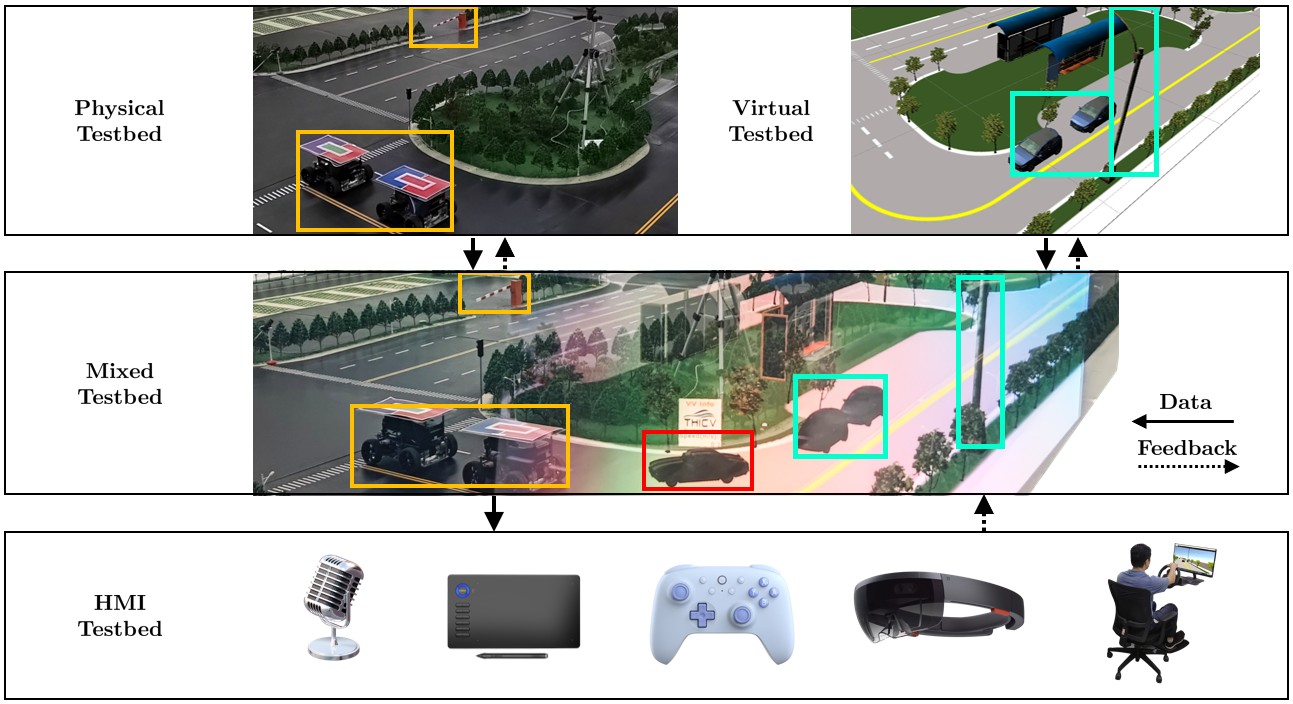}
	%\vspace{-2mm}
	\caption{Schematic of I-VIT, IMPACT's practical implementation. In the mixed testbed, the physical vehicles and infrastructure in the physical testbed (outlined in gold), the virtual vehicles and infrastructure in the virtual testbed (outlined in cyan), the virtual vehicle in the virtual environment of a driving simulator (outlined in red) and human operators in the HMI testbed operate synchronously and interact seamlessly across environments in real time. 
    Consequently, I-VIT is an interactive VICS-DT testing environment, where human operators can directly interact with the physical and virtual entities in I-VIT. The demonstration videos are available at our project website: \url{https://dongjh20.github.io/IMPACT}.
    }
	\label{fig-I-VIT-overview}
	\vspace{-2mm}
\end{figure}

\subsection{Contributions}

\begin{figure*}[t]
	%\vspace{-3mm}
	\centering
	\includegraphics[width=0.99\textwidth]{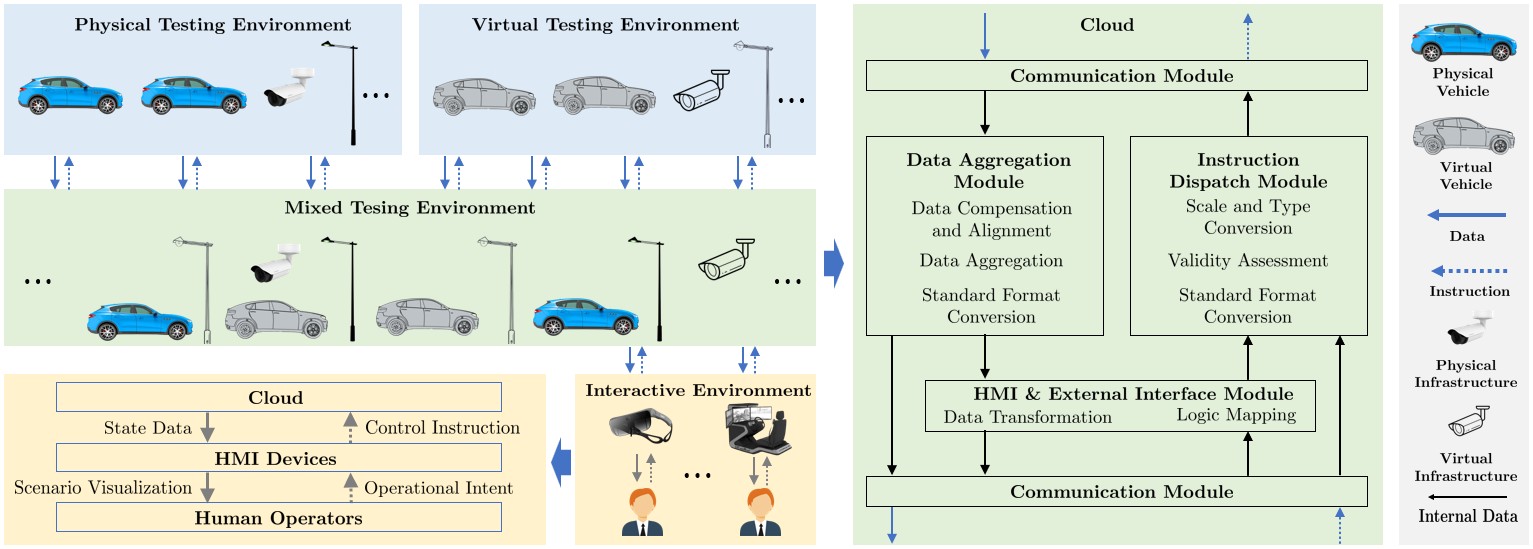}
	\vspace{-2mm}
	\caption{System architecture of the IMPACT framework, consisting of the physical, virtual, mixed testing environments and the interactive environment.}
	\label{fig-IMPACT-framework}
	\vspace{-2mm}
\end{figure*}

Based on the practical requirements of DT-based VICS testing and the novel mixedDT framework, this paper proposes IMPACT, the Interactive Mixed-digital-twin Paradigm for Advanced Cooperative vehicle-infrastructure Testing, an interactive VICS testing framework. 
As illustrated in Fig.~\ref{fig-IMPACT-framework}, IMPACT comprises physical, virtual, and mixed testing environments, alongside an interactive environment. 
The mixed testing environment integrates the other three environments into a unified environment, enabling entities to operate synchronously and interact seamlessly across environments in real time.
Notably, human operators can directly interact with both physical and virtual entities, thereby stochastically influencing their behaviors and states. 
We further develop I-VIT (Interactive Vehicle-Infrastructure Testbed) to actualize the IMPACT framework, as depicted in Fig.~\ref{fig-I-VIT-overview}. 
The main contributions of this paper are summarized as follows:

\begin{itemize}
    	\item Beyond ``Optimizable'', we firstly propose the ``Interactable'' level in VICS-DT, and further present the IMPACT framework for DT-based VICS testing. By directly incorporating the highly uncertain and unpredictable human behaviors to generate corner cases in the mixed testing environment, IMPACT overcomes the inherent limitations of conventional AI-driven generation methods.

	\item By introducing the state-of-the-art mixedDT framework into VICS,  IMPACT empowers virtual entities to exist independently and interactively rather than as typical digital counterparts, thereby enabling action-level interaction between physical and virtual entities, namely ``Physical-Virtual Action Interaction''.  This facilitates safe VICS testing under corner cases involving physical environments and entities, rather than purely in simulation.
	
	\item We implement the IMPACT framework by developing the I-VIT experimental platform. Case studies, including corner cases generated through human interactions and physical-virtual integration tests, demonstrate the effectiveness of both the platform and the proposed  framework.
\end{itemize}

The rest of this paper is organized as follows. 
Section~\ref{sec.2} presents the IMPACT framework, and Section~\ref{sec.3} describes its practical implementation, the I-VIT experimental platform. Section~\ref{sec.4} presents the case studies conducted on I-VIT, and Section~\ref{sec.5} concludes this paper.

\section{The IMPACT Framework}
\label{sec.2}

In this section, we present the overall architecture and methodology of the proposed IMPACT framework. We first provide an overview of the framework and then elaborate each of its components in detail.

\subsection{Framework Overview}
\label{subsec-framework overview}

As shown in Fig.~\ref{fig-IMPACT-framework}, IMPACT consists of four parts: 
\begin{enumerate}
 \setlength\itemsep{0.1em}
    \item Physical testing environment, which primarily comprises physical vehicles and roadside infrastructure, including sensors such as cameras and radars, as well as roadside facilities such as streetlights, traffic signals, and barrier gates. The physical environment inherently involves noise and disturbances and serves as the ultimate testing and operational environment for VICS.
    \item Virtual testing environment, corresponding to the physical testing environment, comprises virtual vehicles and virtual roadside infrastructure, including sensors such as cameras as well as roadside facilities such as streetlights. Unlike the physical environment, the virtual environment offers high flexibility and configurability for VICS testing. 
    
    \item Interactive environment, which broadly includes dedicated interfaces such as Mixed Reality head-mounted displays (MR HMDs) and  driving simulators, as well as general interfaces such as monitors, microphones, drawing tablets, keyboards, and mouses. The interactive environment captures human operational intent and converts it into transferable control signals, serving as the human-intent input interfaces and enabling
    interactive testing of VICS. 
    
    \item Mixed testing environment, which comprises the communication, data aggregation, instruction dispatch, and human-machine interface (HMI) \& external interface modules. It runs in the cloud and serves as the core of IMPACT and the key enabler of interactive VICS testing. It bridges the physical testing environment, the virtual testing environment, and the interactive environment into a unified environment, where they operate synchronously and interact seamlessly across environments in real time.
\end{enumerate}

In the following, we elaborate on each part.

\subsection{Physical Testing Environment}
\label{subsec-physical environment}

The physical testing environment is the environment in which VICS is ultimately deployed and operated, and it also serves as the environment for pre-deployment testing. 
Within the physical testing environment, the most critical aspects regarding interactivity are the precise state acquisition of physical entities and the proper design of control interfaces. We elaborate on this by analyzing the two primary components of VICS, vehicles and roadside infrastructure, which may require distinct techniques and strategies.

For physical vehicles, the key states of interest are primarily their kinematic properties, such as position, heading, speed, and acceleration. Accurate acquisition of vehicle states is relatively straightforward and can be obtained in real time from onboard sensors (e.g., GNSS). Methods for delay compensation and state estimation has also been well established in related research. 
Consequently, digital twinning of vehicle states has constituted a core component in previous VICS-DT. 
Vehicle control interfaces possess a hierarchical structure and can be broadly categorized into four levels: (1) the Trajectory or Path Level, which provides target waypoints; (2) the Kinematic Level, specifying target velocity and target steering angle; (3) the Dynamic Level, dictating target acceleration alongside target steering angle (or target yaw rate); and (4) the Actuator Level, involving throttle opening (pedal stroke), braking pressure, and steering torque. Regarding externally exposed control interfaces, the first two levels are generally sufficient. The dynamic and actuator levels should remain strictly localized within the vehicle's native control system to ensure operational safety and maintain traceability of external control inputs.
The mature autonomous driving architectures have also typically focused on the design of the first two levels, given that the latter two levels are already highly mature and reliable.

Regarding roadside infrastructure, we consider two categories: sensors and facilities. 
Sensors merely acquire and input data into the system; as non-interactive entities, they are not discussed further. 
For roadside facilities, the states of streetlights, traffic signals, and barrier gates are discrete and take only a limited number of values.
These facilities are typically equipped with mature local controllers are connected to area-specific control boards.
Consequently, 
their accurate states can be directly obtained from the corresponding control boards and transmitted via Roadside Units (RSUs).
In terms of control interfaces, it is sufficient to interact with the area-specific control boards, rather than directly interfacing with the local controllers of individual roadside facilities.
For instance, a standard approach involves transmitting specific bit-level control commands to the control boards to control the on/off states of specific streetlights. 
This requires only access to the control boards’ bit-level command mapping table.
Thus, the control of roadside facilities is considerably simpler and more straightforward.
Compared to vehicle interaction, this constitutes an inherently safer mode of interaction within the physical testing environment.

\subsection{Virtual Testing Environment}
\label{subsec-virtual environment}

The virtual testing environment is essentially a mature simulation environment, in which state acquisition and control of virtual vehicles and virtual roadside facilities are straightforward and can be realized by directly invoking the corresponding APIs. The control interfaces discussed within the context of the physical testing environment are also applicable in the virtual testing environment.

Unlike conventional VICS-DT systems built on classical DT paradigms, the mixedDT-based IMPACT framework allows virtual entities to exist independently, rather than being limited to digital counterparts of physical entities.
This characteristic provides the underlying  architectural foundation for direct interactions between physical and virtual entities. Virtual entities are no longer confined to passively receiving information from physical entities and providing optimized feedback; instead, they exist as independent entities, allowing direct interaction with physical entities.
Naturally, they can also simply be utilized as twin entities for physical entities, namely their digital counterparts.

Therefore, although the virtual testing environment mirrors the physical testing environment in terms of its components, the attributes of these components can differ substantially. 
Consequently, the virtual testing environment offers significantly greater flexibility and scalability than the physical testing environment: the number of vehicles, their behaviors, and their low-level dynamic characteristics can be arbitrarily designed; the number of sensors and their performance parameters can be configured on demand; and the control policies for roadside facilities (e.g., streetlights) can be arbitrarily adjusted and tested without incurring real energy consumption. 
Such capabilities are impractical in the physical testing environment, or would require substantial resources and lead to significant waste.

Furthermore, the interactivity of these highly configurable virtual entities facilitates a variety of promising VICS testing applications. 
For example, when testing smart streetlights (which dynamically illuminate areas ahead and extinguish those behind), coupling physical vehicles with virtual streetlights significantly reduces energy consumption caused by frequent switching. For smart barrier gates (pre-authentication for non-stop passage), preliminary debugging can pair physical vehicles with virtual gates—or vice versa—to prevent physical collision damage due to improper parameter settings.
Naturally, the virtual testing environment also has inherent limitations: the noise and disturbances in the physical environment are difficult to reproduce in simulation, as they are hard to model and predict.

\begin{figure*}[t]
	\centering
	\subfigure[Mechanism for scenario visualization via HMI devices]
	{\includegraphics[width=0.5\textwidth]{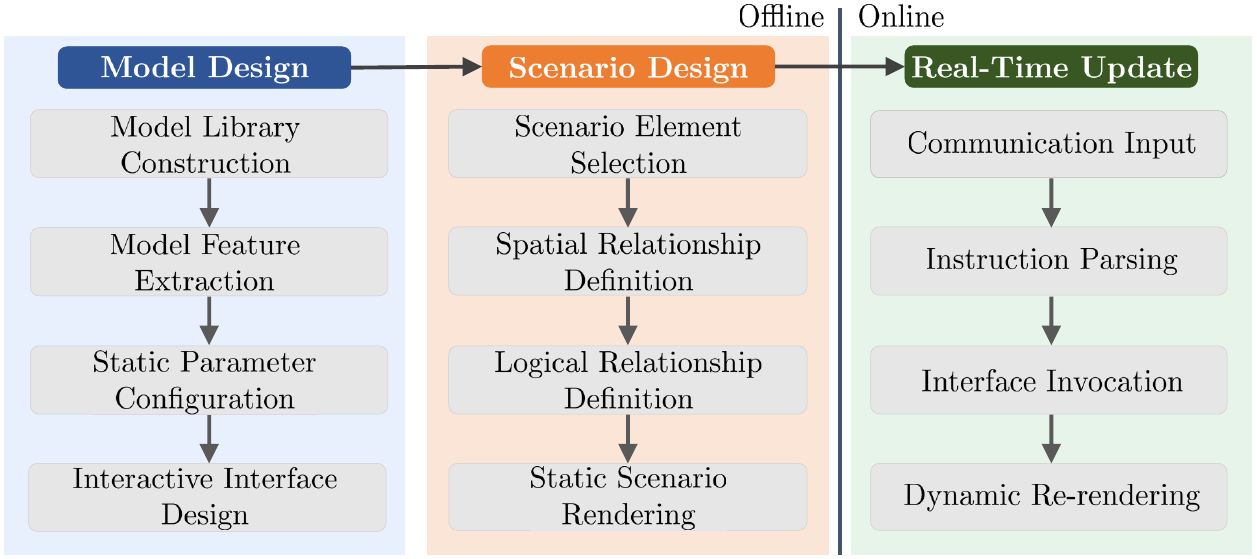}
		\label{fig-HMI-visualization-method}}
	%\hspace{-1mm}
    \subfigure[Scenario visualization by MR HMD]
	{\includegraphics[width=0.45\textwidth]{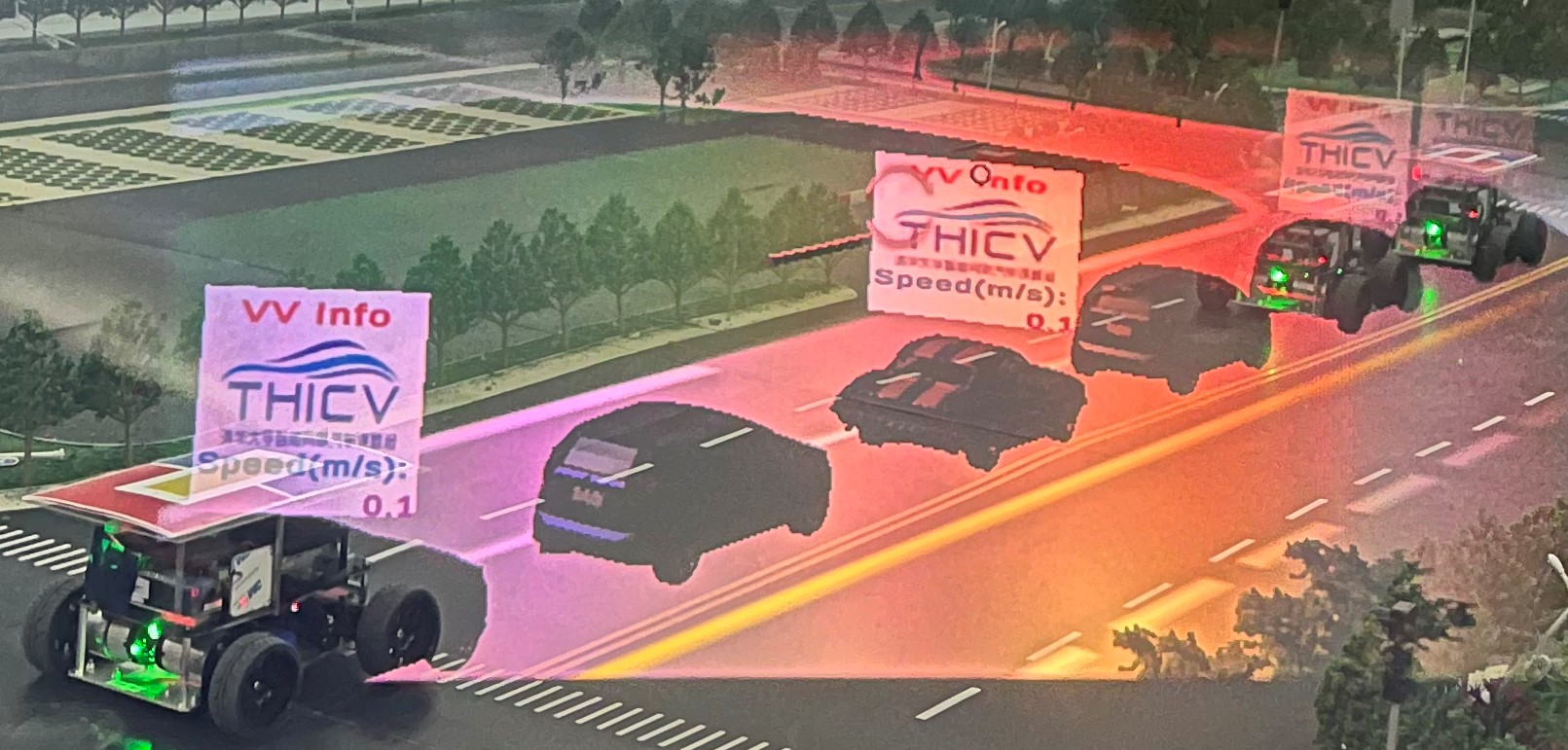}
		\label{fig-HMI-visualization-effect}}
    
    \subfigure[Mechanism for operational intent processing via HMI devices
]
	{\includegraphics[width=0.5\textwidth]{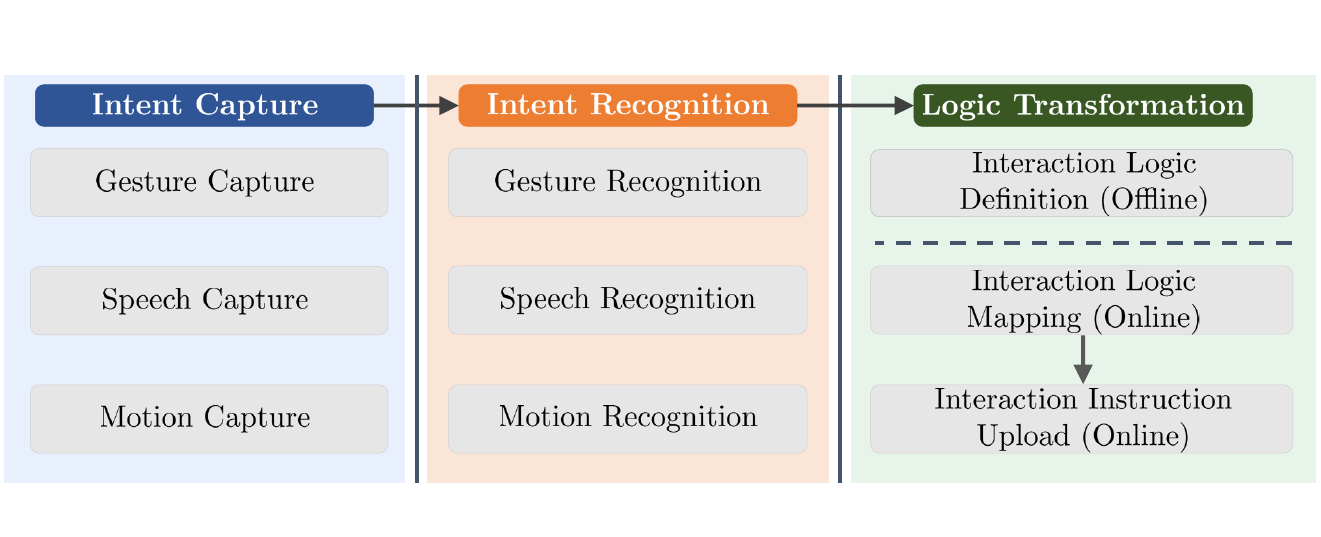}
		\label{fig-HMI-interaction-method}}
    \subfigure[Applying speed perturbations to vehicles via specific gesture]
	{\includegraphics[width=0.45\textwidth]{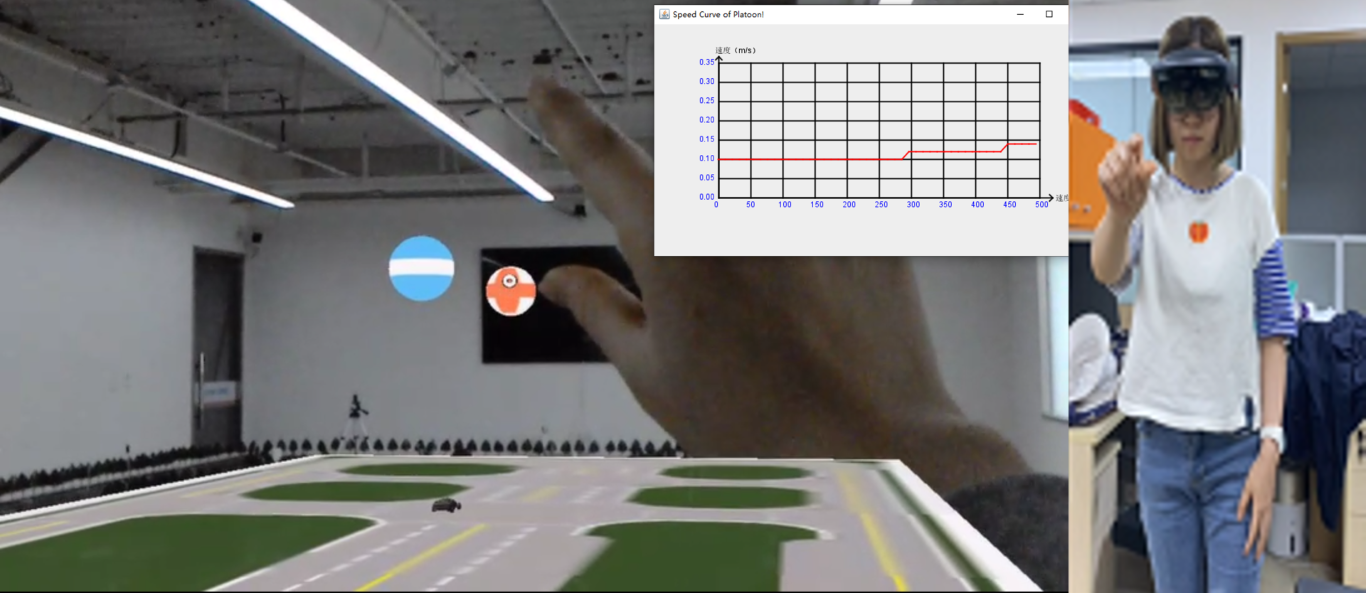}
		\label{fig-HMI-interaction-effect}}
	\caption{Mechanisms for scenario visualization and operational intent processing via HMI devices and their corresponding implementation examples.
	}
	\label{fig-HMI}
\end{figure*}

\subsection{Interactive Environment}
\label{subsec-interactive environment}

The interactive environment primarily serves to capture and convey human operational intent. 
In this environment, human operators use HMI devices to interact directly with physical and virtual entities in the mixed testing environment, thereby introducing highly nonlinear and uncertain human behaviors into VICS testing and naturally generating corner cases.
%This constitutes a distinguishing feature of the IMPACT framework.

The general interactive pipeline is as follows; see the lower-left corner of Fig.~\ref{fig-IMPACT-framework} for illustration. Human operators can observe the real-time operational state of VICS through HMI devices. This is achieved by the HMI devices receiving real-time state updates from the cloud and rendering the scene locally, thereby realizing real-time scenario visualization. Since the entity models are stored on the HMI devices and only state data are transmitted, the communication bandwidth requirement remains modest. Meanwhile, human operators can convey control intentions via HMI devices, for instance, by performing gestures at the gaze point of an MR HMD. 
The HMI devices capture the operator’s intent and, according to predefined interaction logic, translate it into specific control instructions; for instance, the gesture may be converted into a sinusoidal speed perturbation applied to a specific vehicle. The HMI devices then package the instruction in a predefined message format and transmit it to the cloud. The cloud subsequently validates the command (e.g., whether it would result in a negative speed), maps it to the corresponding control inputs for the target physical or virtual entity, and dispatches the resulting commands to it for execution. 
% Ultimately, this pipeline enables observation and evaluation of the VICS's response to sporadic and abrupt speed fluctuations, thereby facilitating performance testing under corner-case conditions.
The detailed implementation mechanisms for scenario visualization and operational intent processing are illustrated in Figs.~\ref{fig-HMI-visualization-method} and~\ref{fig-HMI-interaction-method}, while their corresponding implementation examples are shown in Figs.~\ref{fig-HMI-visualization-effect} and~\ref{fig-HMI-interaction-effect}. This will be further elaborated with concrete examples in Section~\ref{subsec-HMI-testbed}.

Moving beyond classical DT-based VICS testing, which primarily focuses on data-driven optimization and VICS-DT is observed through screens (L4-level VICS-DT), IMPACT offers a new perspective for VICS testing. 
Human operators can directly interact with the VICS-DT system via HMI devices, through either dedicated or general-purpose interfaces. Dedicated HMI devices include MR HMDs, driving simulators, and haptic or force-feedback devices, whereas general-purpose devices encompass microphones, keyboards, mice, game controllers, touchpads, and drawing tablets.
These diverse HMI devices enable capturing human operational intent across multiple modalities and generating a wide range of interactive test cases.

Notably, recent breakthroughs in Multimodal Large Language Models (MLLMs) further lower the barrier to human-machine interaction. 
MLLMs can themselves serve as powerful HMI interfaces, translating unstructured human inputs (e.g., speech and gestures) into structured command representations. 
This capability enables a efficient generation of diverse corner cases for VICS testing, representing a promising avenue for future development.

\subsection{Mixed Testing Environment}
\label{subsec-mixed environment}

The mixed testing environment is deployed and operated in the cloud. Precisely, within the context of a multi-layer cloud architecture, it resides at the edge-cloud server layer to minimize communication latency as much as possible. 
As illustrated on the left side of Fig.~\ref{fig-IMPACT-framework}, human operators interact with diverse entities from the physical and virtual testing environments via this cloud-based relay. 
Therefore, the mixed testing environment serves as the information hub and control center of the IMPACT framework, namely the entire VICS-DT testing system. 
It mainly comprises four modules: (1) communication module, (2) data aggregation module, (3) HMI \& external interface module, and (4) instruction dispatch module.
Fig.~\ref{fig-HMI-visualization-effect} presents the visualization of the mixed testing environment through the MR HMD. The virtual scene rendered onto the lens is overlaid with the physical scene observed through the lens, providing an intuitive illustration of the mixed testing environment.

The cloud operational scheme proceeds as follows. The communication module continuously receives state data from physical and virtual vehicles and roadside facilities in the physical and virtual testing environments and forwards them to the data aggregation module. In this module, latency compensation and spatio-temporal alignment are first performed, particularly for multi-source measurements of the same entity (e.g., a physical vehicle’s state may be reported by onboard sensors and also inferred by roadside sensing). 
Particularly, given the heterogeneous nature of raw state data (e.g., structured ego-vehicle data vs. roadside point clouds/images), we assume, without loss of generality, that raw roadside data is processed locally at the RSUs. Consequently, only structured data is transmitted to the cloud.
Subsequently, data from diverse physical and virtual entities are aligned and aggregated into a unified coordinate system and spatial domain—the mixed testing environment—yielding standardized, structured data packets ready for direct utilization by other modules. These packets are forwarded to the HMI \& external interface module, where they are transformed to satisfy the requirements of target devices prior to transmission.

Concurrently, the HMI \& external interface module processes incoming data from HMI devices. 
Specifically, by referencing a stored interaction logic mapping table, it translates raw HMI inputs (e.g., an MR HMD message dictating a velocity reduction for a target vehicle) into explicit control instructions. These instructions are then routed to the instruction dispatch module. Here, they are first standardized within the mixed testing environment context and subjected to validity assessment (e.g., preventing out-of-bound velocities or abrupt steering jumps). Finally, utilizing a stored entity-instruction mapping table, the instructions are adapted to the scale and type of the target entity's native environment, and dispatched by the communication module for execution.

As illustrated by the operational scheme, 
diverse and heterogeneous entities (vehicles and infrastructure) from the physical and virtual testing environments, alongside HMI devices from the interactive environment, are integrated into the unified mixed testing environment, where they operate synchronously and interact seamlessly across environments in real time, successfully overcoming the ``organizational siloing” challenge~\cite{grieves2017digital} of classical DT.
Consequently, direct action interactions are established between physical and virtual entities, particularly among vehicles. For instance, a braking maneuver by a physical vehicle can directly trigger the braking of a trailing virtual vehicle. 
This represents the successful realization of ``Physical-Virtual Action Interaction'', moving  beyond conventional data-level exchange, namely ``Physical-Virtual Data Interaction''.

\begin{figure*}[t]
	\centering
	\subfigure[Component and data interaction in IMPACT]
	{\includegraphics[width=0.5\textwidth]{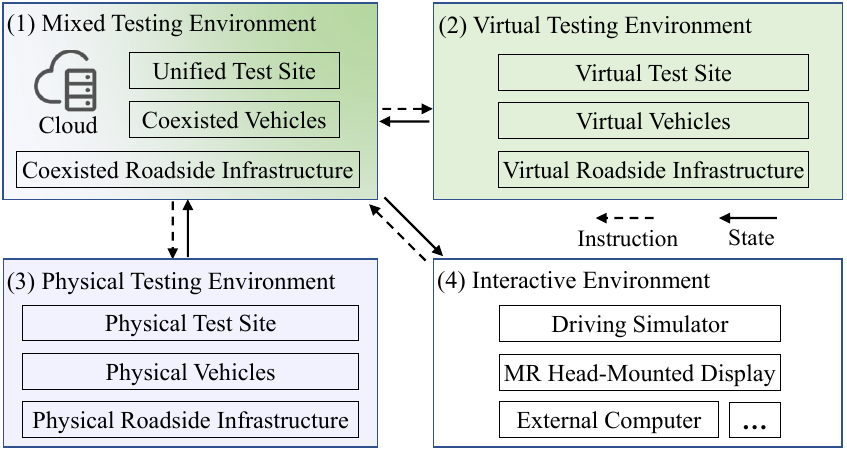}
		\label{fig-IMPACT-flow}}
	%\hspace{-1mm}
    \subfigure[Key communication links in IMPACT's implementation, I-VIT.]
	{\includegraphics[width=0.45\textwidth]{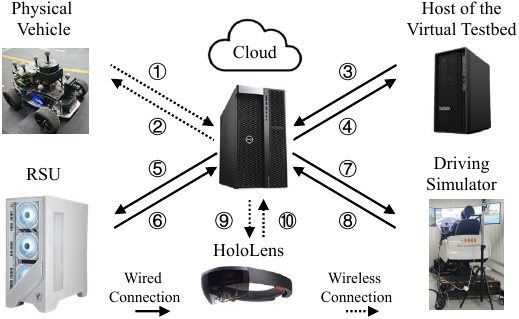}
		\label{fig-communication-link}}
	\caption{Illustration of data interactions within the IMPACT framework and its implementation, the I-VIT platform.
	}
	\label{fig-mixed-environment}
\end{figure*}

Moreover, the modular design of the mixed testing environment also enables the instruction-execution decoupling within the IMPACT framework. 
This decoupling introduces significant flexibility and scalability, serving as one cornerstone of interactive VICS testing.
Accordingly, human operators and diverse VICS entities can interact via cloud-based relays; see Fig.~\ref{fig-IMPACT-flow} for illustration. 
On the one hand, the native control interfaces of individual entities remain unexposed, and they interact exclusively with the cloud, thereby ensuring safety. 
On the other hand, the cloud provides straightforward interfaces for HMI devices by leveraging pre-configured interaction logic mapping tables.
This allows the HMI devices to focus on capturing and transmitting human intent, conveniently bypassing the complex and heterogeneous VICS control interfaces.
% Ultimately,  achieving synchronized execution and cross-platform interaction.

Consequently, within the IMPACT framework, highly uncertain and unpredictable human behaviors can be directly mapped to the actions of specific VICS entities, providing valuable corner cases beyond  AI-driven generation.
Several examples are listed below: via MR HMDs, operators can abruptly spawn obstacles at their gaze points, apply velocity perturbations to specific vehicles or trigger their breakdowns, inject specific noise into (or disable) roadside sensors, and actuate barrier gates; via driving simulators, stochastic human driving behaviors can be directly introduced; via game controllers, irregular velocity fluctuations can be applied to specific vehicles; and via drawing tablets, custom, hand-drawn velocity profiles can be injected. 
Collectively, these multimodal interactions effectively and efficiently generate high-quality corner cases for VICS testing.

Furthermore, since the interactive VICS entities can be either physical or virtual, the ``Physical-Virtual Action Interaction'' versatility, coupled with the instruction-execution decoupling, further provides substantial experimental flexibility and scalability, facilitating VICS tests that cannot be achieved by simple platform stacking.
For instance, in smart streetlights testing, the vehicle can be physical and controlled by a human via a driving simulator, directly introducing stochastic human driving behaviors alongside real-world vehicle state update rates and precision, while the streetlights can be virtual to avoid energy costs caused by frequent switching and to support early debugging before the physical streetlights are deployed in VICS.

\begin{figure*}[t!]
	%\vspace{-3mm}
	\centering
	\includegraphics[width=0.99\textwidth]{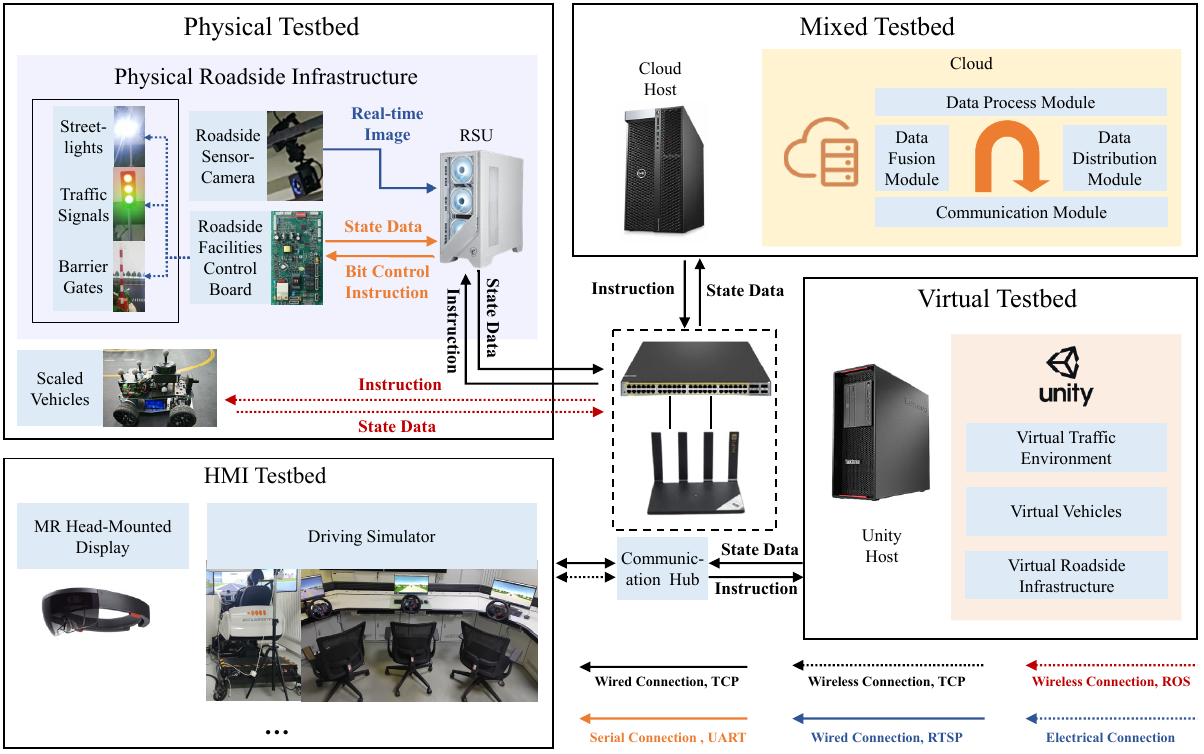}%PDF:0.72
	%\vspace{-2mm}
	\caption{Illustration of the I-VIT architecture and interactions among its four components: physical, virtual, HMI, and mixed testbeds. 
    }
	\label{fig-I-VIT}
	\vspace{-2mm}
\end{figure*}

\begin{figure*}[t!]
	\centering
	\subfigure[Physical traffic environment]
	{\includegraphics[height=0.2\textheight]{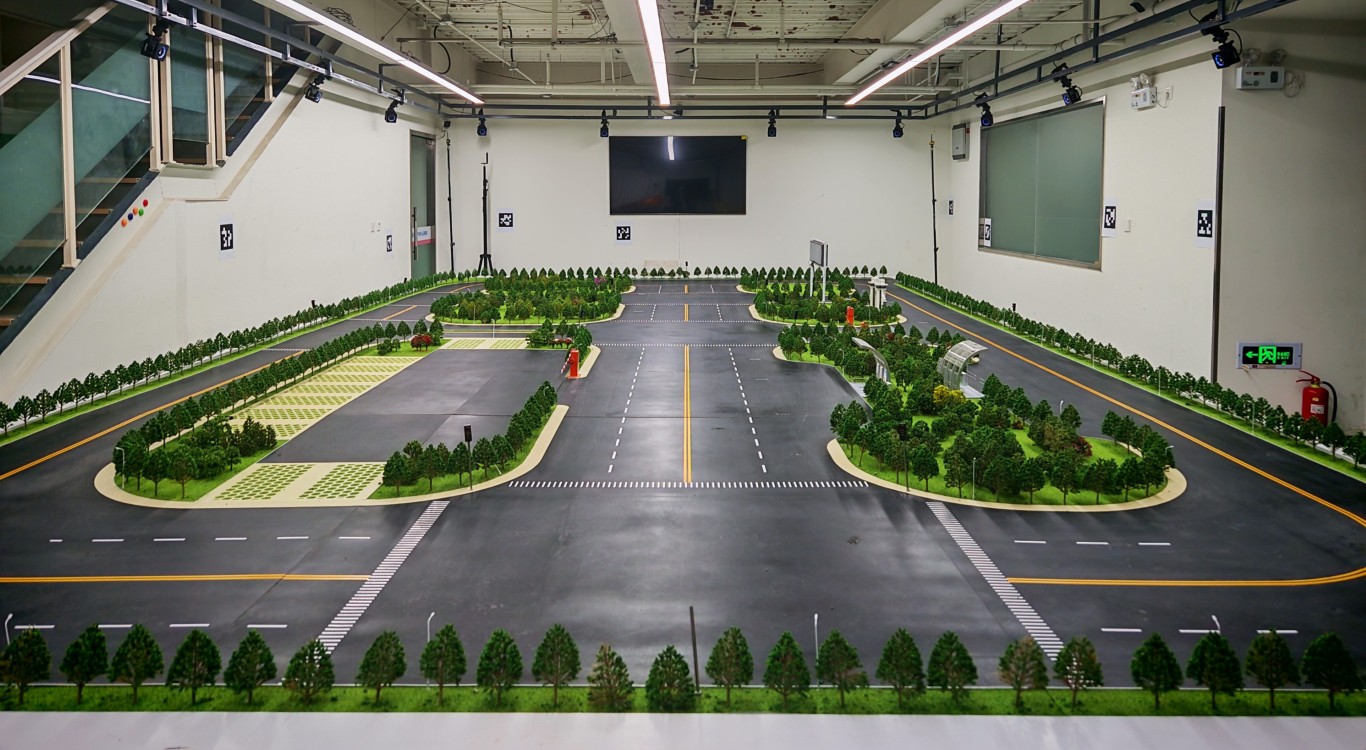}
		\label{fig-physical-testbed}}
	\hspace{3mm}
    \subfigure[Virtual traffic environment]
	{\includegraphics[height=0.2\textheight]{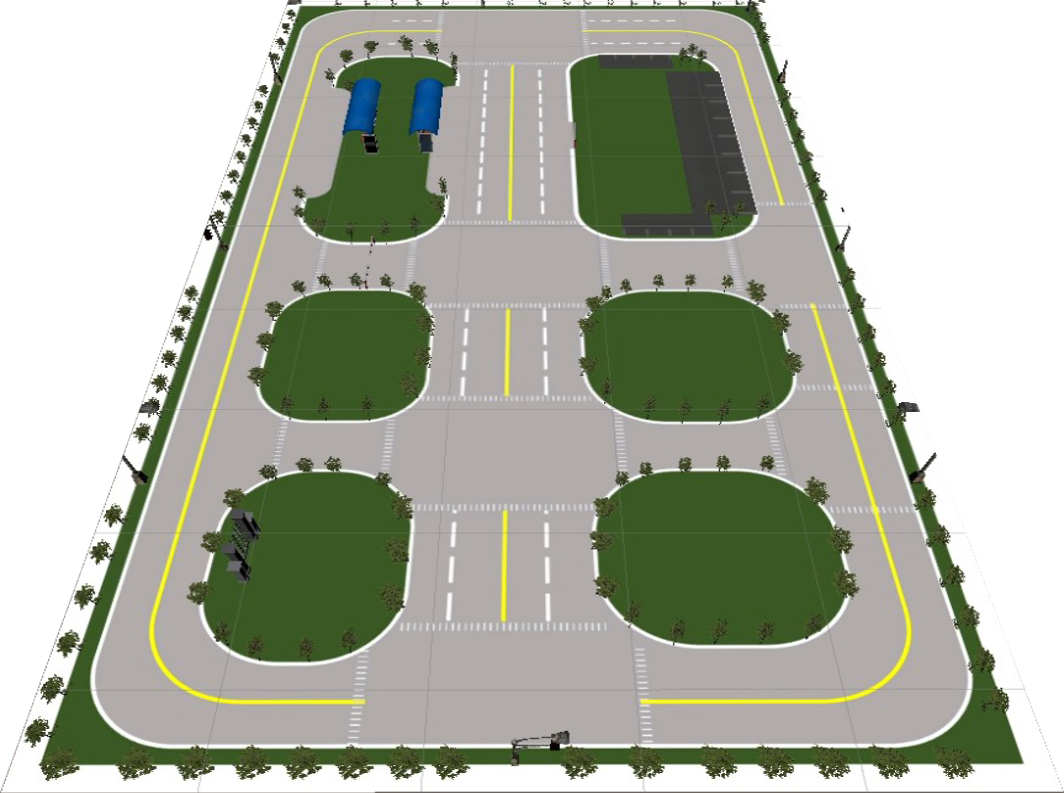}
		\label{fig-virtual-testbed}}
	\subfigure[Serial channel-facility mapping table for physical roadside facilities]
	{\includegraphics[height=0.192\textheight]{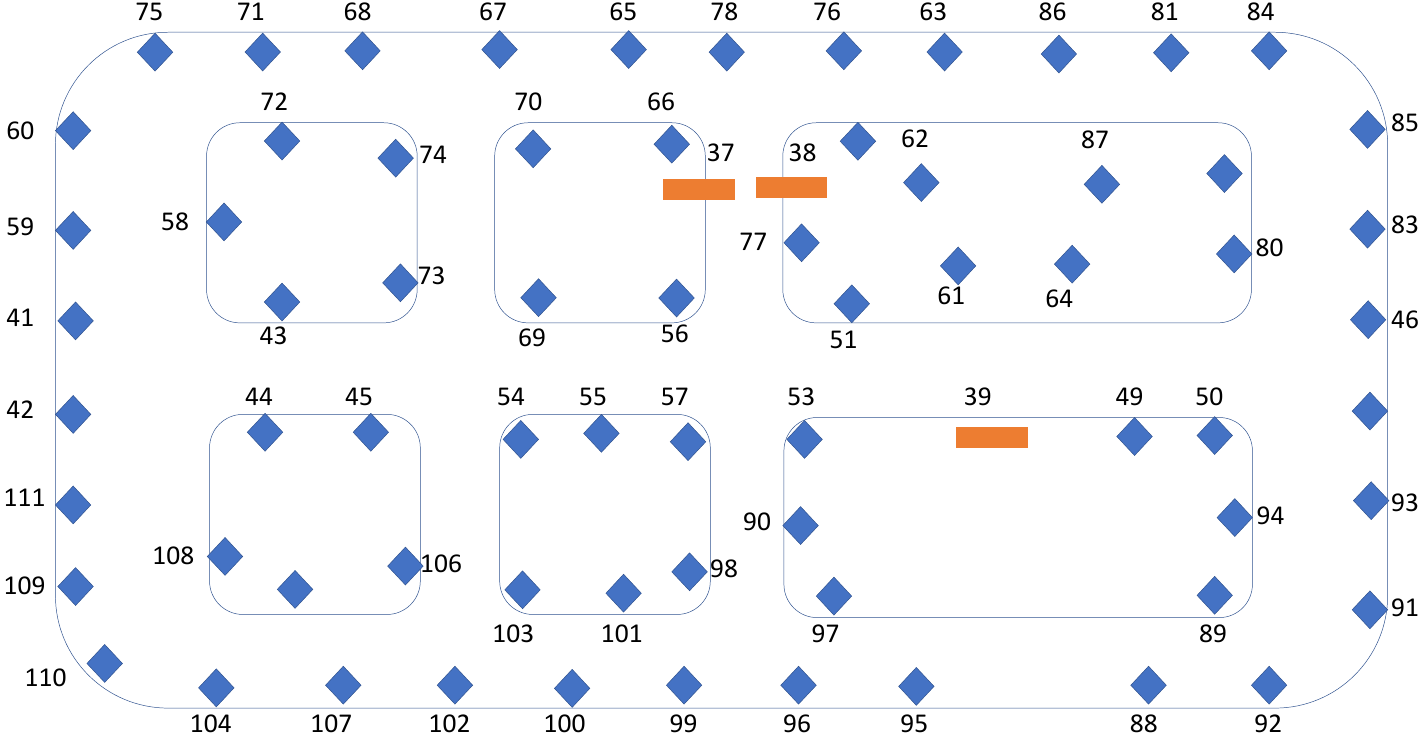}
		\label{fig-physical-mapping-table}}
        \hspace{3mm}
    \subfigure[Physical and virtual vehicles and facilities]
	{\includegraphics[height=0.192\textheight]{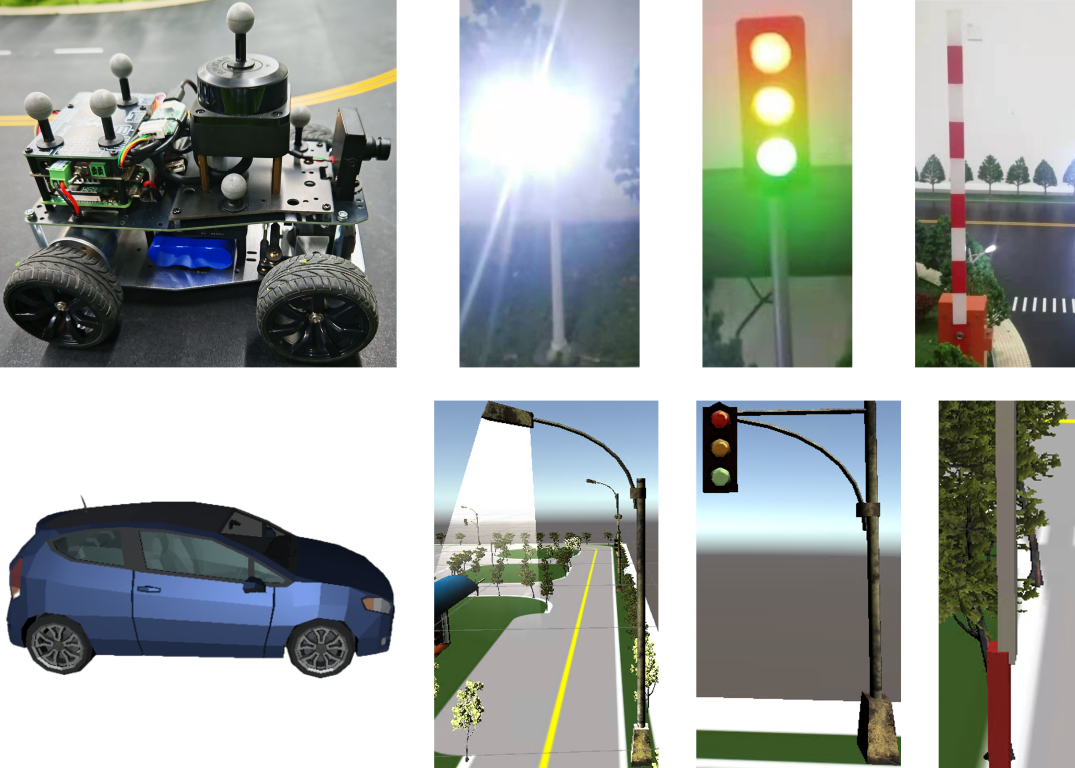}
		\label{fig-vehicle-facilities}}
	\vspace{-2mm}
	\caption{Physical and virtual traffic environments, vehicles and facilities. (a) The physical traffic environment, having a size of $9\,\mathrm{m} \times 5\,\mathrm{m}$. (b) The virtual traffic environment, developed based on the Unity game engine. (c) The serial channel-facility mapping table for physical roadside facilities. Blue denotes streetlights and traffic signals, whereas orange denotes barrier gates. (d) Physical and virtual vehicles and roadside facilities. 
	}
	\label{fig-testbed}
\end{figure*}

\begin{table*}[t!]
\footnotesize
\caption{I-VIT Setups}
\label{tab-setup}
\centering
		\begin{threeparttable}
		\setlength{\tabcolsep}{10mm}{
			\begin{tabular}{cc}
   \toprule
				Component & Setup \\\hline
				Physical sand table & $9\,\mathrm{m} \times 5\,\mathrm{m}$,  $1:14$ with respect to the real-world driving environment \\
                Roadside sensors in the physical testbed & 16 identical cameras surround the physical sand table\\
                Roadside facilities in the physical testbed & $73$ streetlights, $11$ traffic signals and $3$ barrier gates\\
                Scaled physical  vehicles & $215\,\mathrm{mm} \times 190\,\mathrm{mm} \times 125\,\mathrm{mm}$, powered by Raspberry Pi 4B
                \\
                Host of the RSU & CPU: Intel Core i9-14900KF, Memory: 32 GB, GPU: NVIDIA RTX 5080
                \\
                Host of the virtual testbed & CPU: Intel Core i7-10700K, Memory: 16 GB \\
                Host of the G29 simulator & CPU: Intel Core i7-10700, Memory: 16 GB  \\
                Host of the InnoSimulation simulator & CPU: Intel Core i7-8700, Memory: 16 GB  \\
                Edge cloud server for the mixed testbed & CPU: Intel Xeon Gold 5220R, Memory: 64 GB, GPU: NVIDIA RTX 3090\\
                
				\bottomrule
		\end{tabular}}
  
		\end{threeparttable}
	\vspace{-2mm}
\end{table*}

\section{Framework Implementation}
\label{sec.3}

In this section, we present the practical implementation of the IMPACT framework. 
As shown in Figs.~\ref{fig-IMPACT-framework} and~\ref{fig-IMPACT-flow}, the IMPACT framework has a concise and intuitive architecture, consisting of the physical, virtual, and mixed testing environments, alongside the interactive environment. 
Specifically, Fig.~\ref{fig-IMPACT-flow} essentially illustrates the theoretical architecture of an experimental platform based on the IMPACT framework.
Guided by this architecture, our implemented I-VIT naturally comprises four corresponding components—physical, virtual, mixed, and HMI testbeds. 
Fig.~\ref{fig-I-VIT} provides a detailed illustration of the I-VIT architecture and the interaction mechanisms among its components.
Some key parameters of I-VIT are summarized in Table~\ref{tab-setup}.
The implementation details of each component are sequentially elaborated below.

\subsection{Physical Testbed}
\label{subsec-physical-testbed
}

The physical testbed represents the physical testing environment in IMPACT, and consists of a physical traffic environment, physical vehicles and physical roadside infrastructure. 
As shown in Fig.~\ref{fig-physical-testbed}, the physical traffic environment is an affordable, convenient, and repeatable scaled sand table ($1:14$ scaling ratio) that captures the key structure of a road network. Accordingly, the physical vehicles are the scaled ones shown in Fig.~\ref{fig-vehicle-facilities}. 
Their precise states are obtained as follows: markers are attached to the vehicles to enable the FZMotion motion capture system to accurately estimate their poses, with a position accuracy of $0.1 \;\mathrm{mm}$ and an angular accuracy of $0.1^{\circ}$; the real-time vehicle speed is measured using onboard Hall sensors. 
For the control interface, we adopt a mature command pair—target speed and target front-wheel steering angle—which is sent from the cloud via Robot Operating System (ROS) messages.
%, supporting instruction-execution decoupling.

Physical roadside infrastructure is centrally managed by the RSU. It encompasses roadside sensors (specifically, 16 cameras) and physical facilities (including streetlights, traffic signals, and barrier gates); see Fig.~\ref{fig-vehicle-facilities} for illustration. 
These facilities are uniformly governed by a dedicated roadside facility control board.
For facility state acquisition, the RSU periodically sends a query command to the control board via a serial interface. The control board maintains a status bit for each channel and, upon receiving the query, returns a data frame containing the states of all channels. As shown in Fig.~\ref{fig-physical-mapping-table}, the RSU stores a serial channel-facility mapping table; by parsing the returned frame, it obtains the states of all facilities and forwards them to the cloud.
To illustrate the cloud-based control scheme, consider the case where the cloud instructs the RSU to open the left-side barrier gate. The RSU first looks up the mapping table and identifies that the left-side gate corresponds to channel $39$. It queries the current state of this channel and, if the gate is closed, further consults the channel-bit control command mapping table to obtain the open and close commands ($\mathrm{0x26}$ and $\mathrm{0x8A}$, respectively). The RSU then sends $\mathrm{0x26}$ to the control board, and the board actuates the left-side barrier gate accordingly.

\subsection {Virtual Testbed}
\label{subsec-virtual-testbed
}

The virtual testbed corresponds to the virtual testing environment in the IMPACT framework and includes a virtual traffic environment, virtual vehicles, and virtual roadside infrastructure. It is developed using the Unity game engine and thus provides  high-fidelity physics and extensive development flexibility, running at $50\,\mathrm{FPS}$.
%The entities in the virtual testbed exist independently; however, they can also serve as digital counterparts (i.e., twins) of physical entities.

As shown in Fig.~\ref{fig-virtual-testbed}, the virtual traffic environment shares the same road network topology as the physical one, which facilitates rapid state aggregation and seamless interaction between entities across the two testbeds; however, its geometric scale is aligned with real-world roads rather than the scaled setup. 
The virtual vehicles (Fig.~\ref{fig-vehicle-facilities}) are modeled using a classical kinematic bicycle model~\cite{rajamani2011vehicle}. The roadside infrastructure mirrors those in the physical testbed, including virtual sensors and virtual roadside facilities (see Fig.~\ref{fig-vehicle-facilities} for illustration). 
Leveraging Unity’s parent-child hierarchy, we implement a streetlight manager, a traffic-signal manager, and a barrier-gate manager to centrally manage and control the virtual roadside facilities. 

In addition, for entities that are unavailable in the Unity asset library or do not meet our requirements, we can implement custom models. 
For example, a barrier gate can be represented using two cubes for appearance, with the opening and closing motion realized by rotating the arm cube around a pivot, offering high flexibility.

\subsection{HMI Testbed}
\label{subsec-HMI-testbed}

The HMI testbed represents the interactive environment in the IMPACT framework and mainly consists of HMI devices and human operators. 
In I-VIT, the HMI devices primarily include a MR HMD and four driving simulators. 
Specifically, the MR HMD is the Microsoft HoloLens, while the driving simulators consist of one high-fidelity InnoSimulation simulator and three Logitech G29 simulators; see the lower-left corner of Fig.~\ref{fig-I-VIT} for details. 
Based on the mechanisms presented in Figs.~\ref{fig-HMI-visualization-method} and~\ref{fig-HMI-interaction-method}, their respective  visualization and interaction implementation are detailed below.

\subsubsection{MR HMD}

For the HoloLens used in our testbed, we adopt Unity for scenario development, enabling efficient visualization of the virtual platform, which is also developed by Unity.
Specifically, for scenario visualization, as illustrated in Fig.~\ref{fig-HMI-visualization-method}, we first construct a model library (e.g., vehicles, streetlights, and barrier gates). 
For each model, key features are identified to determine its critical parameters; for example, brightness is a key parameter of a streetlight. Static parameters, such as vehicle mass, are then configured. 
Interaction interfaces are further designed for dynamic entities. 
For instance, a streetlight only requires on/off interfaces rather than specific brightness values.
Subsequently, in scenario design, specific model elements are selected, and their spatial and logical relationships are defined to construct and render the static scene. 
These steps are completed offline. 
During online operation, after receiving state data from the cloud, the system identifies the entities to be updated and their new states, invokes the corresponding interfaces, and re-renders the scenario.
The practical visualization effect of the HoloLens is shown in Fig.~\ref{fig-HMI-visualization-effect}.

For operational intent processing, consider the interaction example shown in Fig.~\ref{fig-HMI-interaction-effect}. Following the processing mechanism illustrated in Fig.~\ref{fig-HMI-interaction-method}, once the operator performs a predefined gesture, the HoloLens immediately captures and recognizes it. 
By referencing the predefined interaction logic, this gesture is mapped to a instruction that increases the test vehicle's speed by one unit. The instruction is then uploaded to the cloud, processed by the cloud as described in Section~\ref{subsec-mixed environment}, and dispatched to the vehicle, resulting in a speed change, as depicted by the speed curve in Fig.~\ref{fig-HMI-interaction-effect}.

\begin{figure*}[t]
	\centering
	\subfigure[The physical driving view]
	{\includegraphics[height=0.15\textheight ]{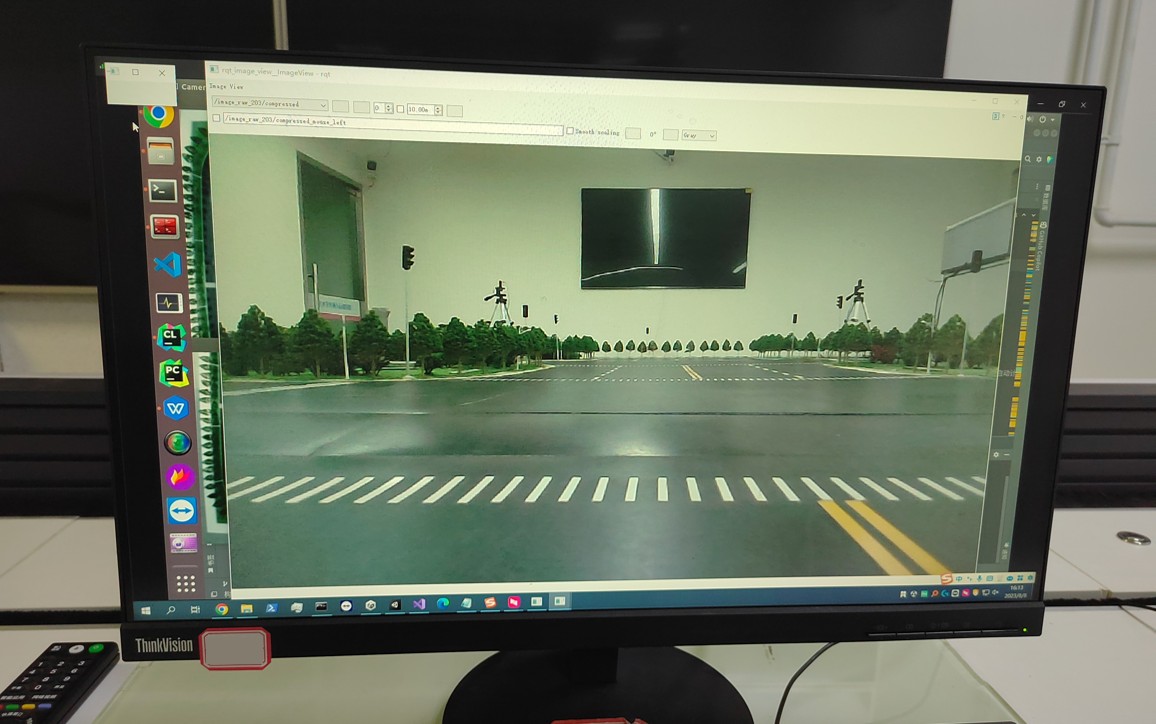}
	\label{fig-physical-view}
	}
	%\hspace{10mm}
	\subfigure[The virtual driving view]
	{\includegraphics[height=0.15\textheight ]{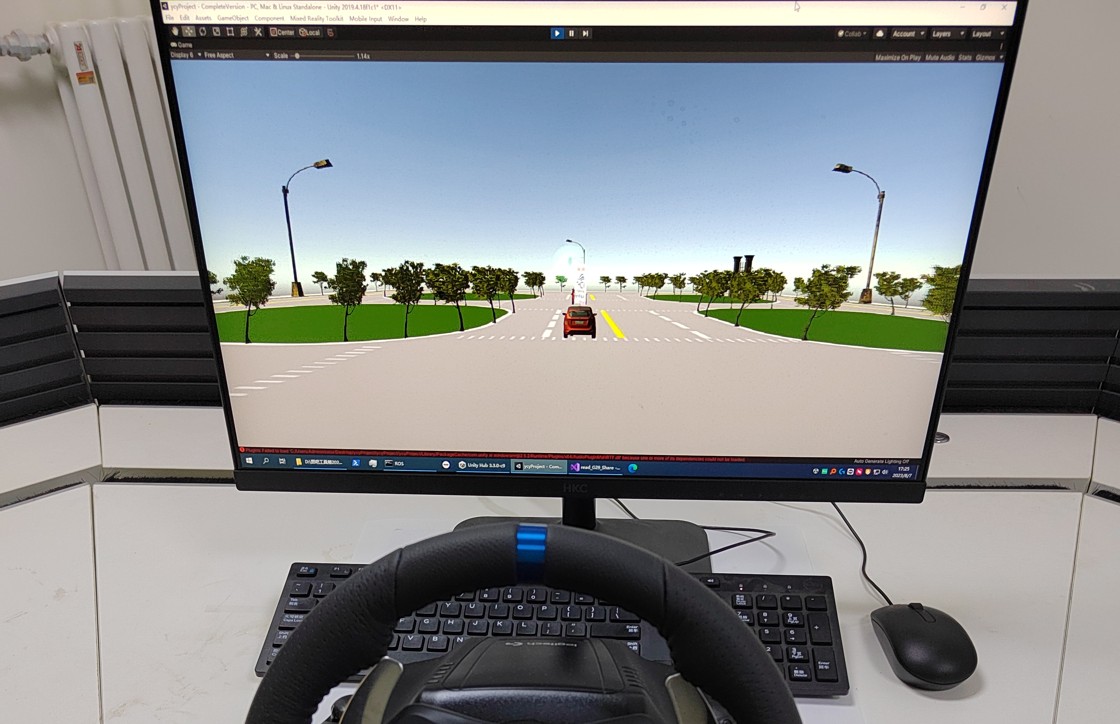}
	\label{fig-virtual-view}
	}
    \subfigure[The information board in the virtual  view]
	{\includegraphics[height=0.15\textheight ]{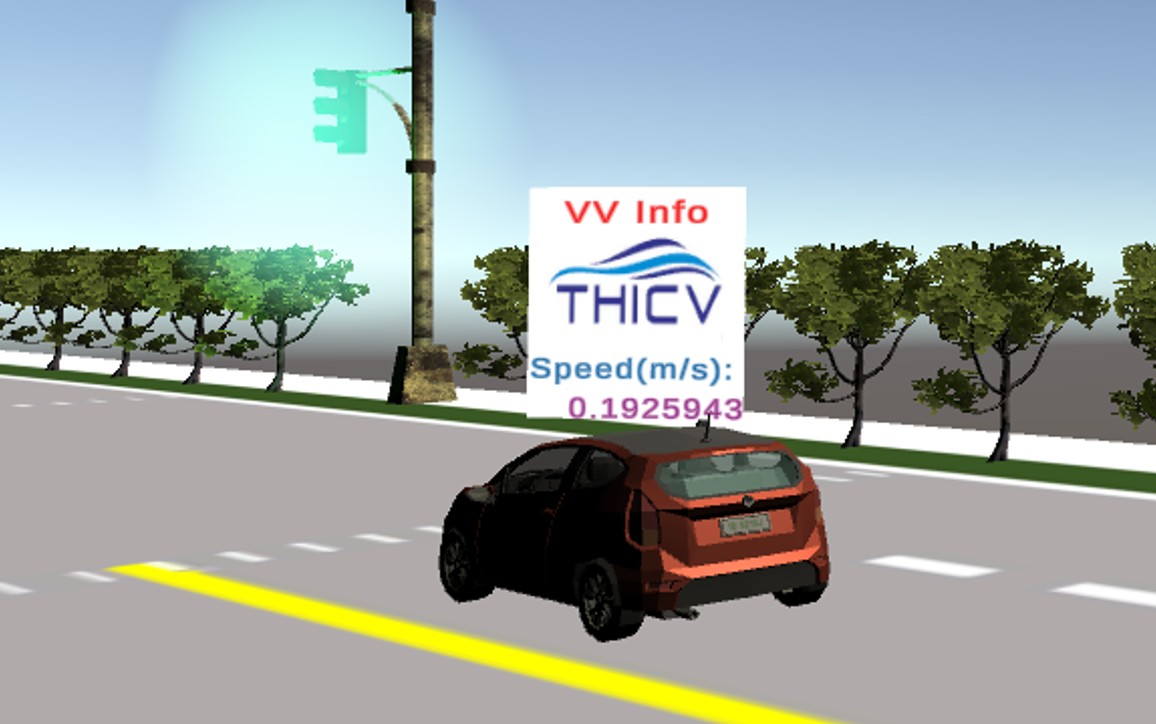}
	\label{fig-speed-promt}
	}
	\vspace{-1mm}
	\caption{Driving view and information board. In (a) and (b), the snapshots of physical and virtual views are presented respectively; see the screens for the specific view. In (c), the real-time speed of the  preceding vehicle in the virtual driving view is displayed on the information board on the top of the preceding vehicle.}
	\label{fig-view-promot}
  \vspace{-4mm}
\end{figure*}

\subsubsection{Driving Simulator}

Driving simulators are mature HMI devices equipped with steering wheels, accelerators, and brake pedals.
Leveraging the IMPACT framework, we integrate them into the VICS-DT testing system to directly manipulate physical or virtual vehicles within it.
Scenario visualization for the simulator is unique, representing a first-person driving view. 
We provide the physical driving view (Fig.~\ref{fig-physical-view}) via the physical vehicle's onboard camera, and the virtual driving view (Fig.~\ref{fig-virtual-view}) via a virtual camera attached to the virtual vehicle. 
Furthermore, as shown in Fig.~\ref{fig-speed-promt}, the highly configurable virtual driving view can be augmented with information boards displaying the speed of the preceding vehicle, facilitating V2X-related tests. 
Notably, enabled by the inherent mixedDT mechanism of IMPACT, when a virtual vehicle is selected as the twin of a physical vehicle, the physical vehicle can be controlled through the virtual driving view of this twin vehicle.

As highly dedicated HMI devices, driving simulators offer extremely direct interfaces. Driver behaviors are directly captured and uploaded to the cloud, processed as presented in Section~\ref{subsec-mixed environment}, and dispatched to the target vehicle for execution, which triggers scenario visualization updates, thereby completing the interaction loop.

\begin{table}[t!]
\footnotesize
	\begin{center}
		\caption{Measurement of Communication Delays in I-VIT (Corresponding to Fig.~\ref{fig-communication-link}, Unit: $\mathrm{ms}$)}\label{tab-commuDelay}
		\begin{tabular}{ccccc}
			\toprule
			\tabincell{c}{Link}   &\tabincell{c}{Gaussian Mean} &\tabincell{c}{Standard Deviation}  &\tabincell{c}{ $99^\mathrm{th}$ Pencentile}  \\\hline
			\circled{1}/\circled{2}     & $1.33$ & $0.66$ & $2.86$ \\
			\circled{3}/\circled{4}     & $0.38$ & $1.17$ & $3.09$ \\
			\circled{5}/\circled{6}     & $1.30$ & $0.57$ & $2.63$ \\
			\circled{7}/\circled{8}     & $0.36$ & $2.74$ & $6.74$ \\
            \circled{9}/\circled{10}     & $4.23$ & $1.72$ & $8.23$ \\
			\bottomrule
            
		\end{tabular}
	\end{center}
	\vspace{-4mm}
\end{table}

\subsection {Mixed Testbed}
\label{subsec-mixed-testbed}

The mixed testbed represents the mixed testing environment within the IMPACT framework and is directly deployed and operated on the edge cloud servers. Guided by the IMPACT framework, it comprises the communication, data fusion, data processing, and data distribution modules, with its operational mechanisms  fully following the paradigms presented in Section~\ref{subsec-mixed environment}. 
By integrating diverse and heterogeneous entities from the physical, virtual, and HMI testbeds into this unified environment, I-VIT enables them to operate synchronously and interact seamlessly across testbeds in real time.

Specifically, within the mixed testbed, human operators can directly interact with and influence both physical and virtual entities in I-VIT, rather than being limited to passive observation. 
Overall, HMI devices (e.g., HoloLens, driving simulators) visualize cloud-transmitted real-time states for operators, whose uploaded operational intentions are then converted by the cloud into state updates of corresponding physical and virtual entities.
The underlying mechanisms and practical significance of such interactive VICS testing have been extensively elaborated in Section~\ref{subsec-mixed environment}, while the implementation details are presented in Section~\ref{subsec-HMI-testbed}.
Such interaction directly introduces unpredictable human behaviors into VICS testing on I-VIT, enabling the generation of corner cases that effectively complement AI-based ones, and thereby facilitating a more comprehensive evaluation of the tested algorithms.

The interactive VICS testing in I-VIT relies on the proper operation of the mixed testbed, which in turn depends on real-time, high-quality communication.
Fig.~\ref{fig-I-VIT} comprehensively illustrates the communication contents and protocols among entities in I-VIT.
The core communication hub consists of a Gigabit Ethernet switch and a router supporting wireless connectivity.
Fig.~\ref{fig-communication-link} summarizes the key communication links in I-VIT, where highly reliable and low-latency wired communication serves as the primary transmission mode. 
The one-way communication delays of these links are presented in Table.~\ref{tab-commuDelay}.
Since the physical traffic environment is scaled, the wireless communication link between the cloud and the physical vehicle still exhibits very low latency. 
The RSU-cloud link shows slightly higher wired latency due to the need to handle high-volume video streams from roadside cameras and a more complex network topology, but it still remains at a very low level.

In addition, the mixed testbed serves as the unified external interface for I-VIT. By communicating with the cloud via predefined formats and protocols, external devices can receive real-time state data of I-VIT entities and return control commands. 
For instance, the high-fidelity InnoSimulation driving simulator in HMI testbed has its own independent virtual traffic environment based on SCANeR Studio. 
Through direct interaction with the cloud, it is integrated into the overall system operation, supporting  the experiment to be presented in Section~\ref{sec.4}. Consequently, this architectural design of the mixed testbed renders the interactive VICS testing within I-VIT highly flexible and scalable.

%% past content
%Consequently, the behaviors of physical and virtual HDVs can truly affect those of physical and virtual CAVs, and vice versa. 
%In other words, four types of vehicles from diverse platforms and environments coexist and interact with each other indirectly via the mixed platform. 

\section{Case Study}
\label{sec.4}

In this section, based on the I-VIT platform, we adopt the classic vehicular platooning system within VICS as a representative case. 
Comprehensive experiments are conducted to demonstrate the capability of I-VIT in conducting interactive VICS testing.
We first detail the experimental setups, followed by a comprehensive analysis of the results.

\begin{figure*}[t!]
	%\vspace{-3mm}
	\centering
	\includegraphics[width=0.9\textwidth]{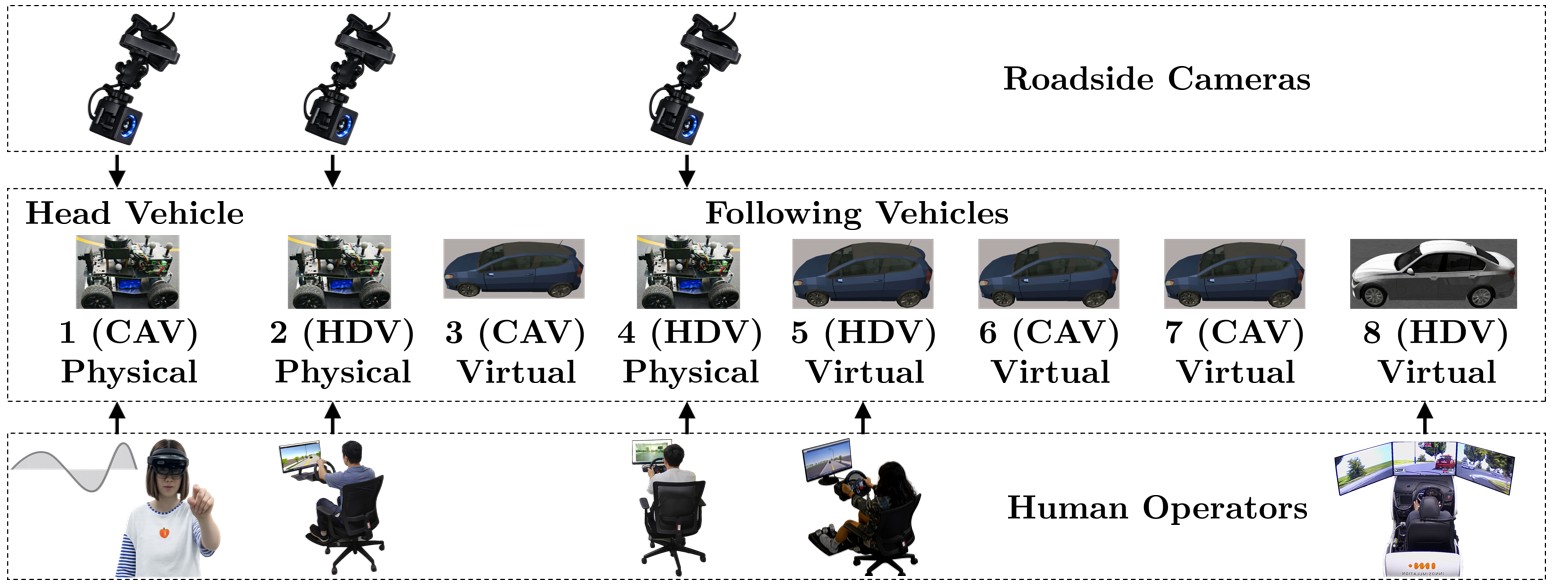}%PDF:0.72
	%\vspace{-2mm}
	\caption{Illustration of the experimental setup. 
    }
	\label{fig-exp-setup}
	\vspace{-2mm}
\end{figure*}

\subsection{Experimental Setup}

The experimental setup is illustrated in Fig~\ref{fig-exp-setup}. 
In I-VIT, the precise states of physical vehicles are obtained via roadside cameras. 
The platoon consists of eight vehicles, including both physical and virtual vehicles. 
In particular, the last virtual vehicle (vehicle $8$) originates from the native virtual environment of the high-fidelity InnoSimulation simulator and is integrated into the experiment via the cloud. 
Human operator interaction  involves two primary aspects:
\begin{itemize}
    \item One operator triggers interventions in a stochastic manner via the MR HMD (HoloLens), imposing a sudden braking perturbation on the head vehicle.
    \item Four operators directly manipulate the physical and virtual vehicles via the simulator interfaces, comprising one high-fidelity InnoSimulation simulator and three Logitech G29 simulators; see the lower part of Fig.~\ref{fig-exp-setup} for illustration.
\end{itemize}

\begin{figure*}[t]
	%\vspace{1mm}
	\centering
	\subfigure[Visualization of the platoon by Unity]
	{\includegraphics[width=0.55\textwidth]{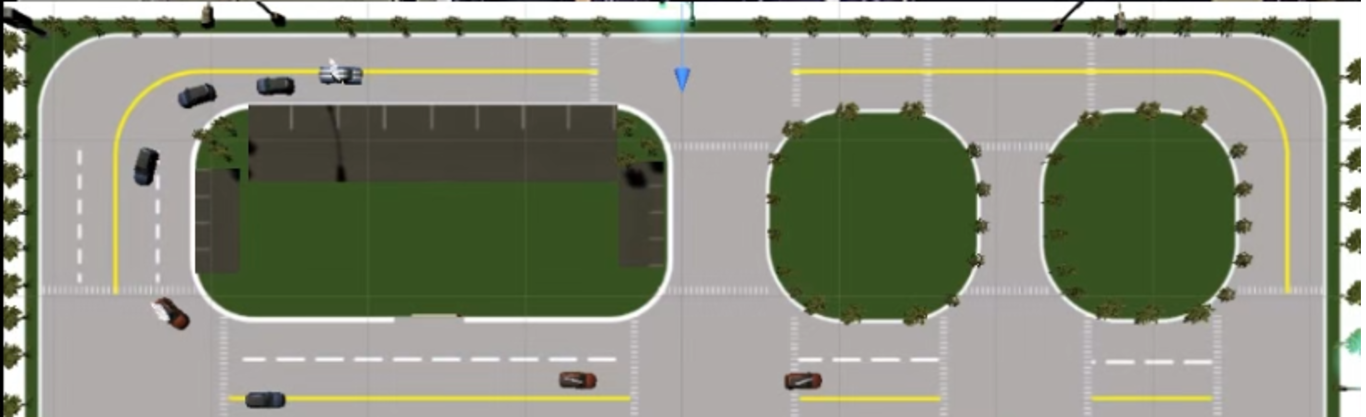}
		\label{fig-exp-snapshot-unity}}
	%\hspace{4mm}
	\subfigure[Control the vehicle via the InnoSimulation simulator]
	{\includegraphics[width=0.41\textwidth]{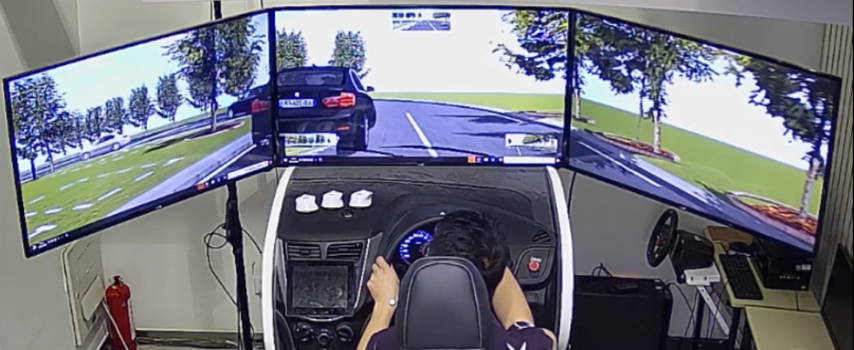}
		\label{fig-exp-snapshot-inno}}
	\vspace{-1mm}
	\caption{Snapshots of the experiment. (a) Visualization of the entire platoon projected into the virtual space by Unity. The platoon configuration is presented in Fig.~\ref{fig-exp-setup}. (b) A human operator controls the last vehicle (Vehicle $8$) via the InnoSimulation simulator. The experimental videos are available at our project website: \url{https://dongjh20.github.io/IMPACT}. 
	}
	\label{fig-exp-snapshot}
\end{figure*}

The remaining vehicles employ the classical Cooperative Adaptive Cruise Control (CACC) algorithm~\cite{milanes2013cooperative} for longitudinal control, and the typical preview trajectory tracking controller~\cite{amer2017modelling} for lateral control.
Consequently, the platoon includes both Connected and Autonomous Vehicles (CAVs) controlled by the CACC algorithm and Human-Driven Vehicles (HDVs).
The detailed composition of the platoon is depicted in the middle part of Fig~\ref{fig-exp-setup}.
Specifically, the platoon travels along the upper half of the physical and virtual testbeds, corresponding to the lanes enclosed by the yellow lines in Fig.~\ref{fig-exp-snapshot-unity}. 
In the absence of external perturbations, the head vehicle maintains a default speed of $10.08\,\mathrm{km/h}$. 
The braking perturbation triggered by human interaction consists of three phases: decelerating to $1.01\,\mathrm{km/h}$ with a deceleration of $0.28\,\mathrm{m/s^2}$, maintaining the low speed for $20$ seconds, and subsequently  returning to $10.08\,\mathrm{km/h}$ after $12$ seconds.
Fig.~\ref{fig-exp-snapshot-unity} captures a snapshot of the experimental process, visualizing the entire platoon projected into the virtual space and allowing observation of its real-time operating states.
The experimental videos are available at our project website: \url{https://dongjh20.github.io/IMPACT}.
%Fig.~\ref{fig-exp-snapshot-inno} depicts the real-time view of the human operator manipulating the vehicle via the driving simulator.

\begin{figure*}[t]
	%\vspace{1mm}
	\centering
	\subfigure[Velocity profiles]
	{\includegraphics[width=0.45\textwidth]{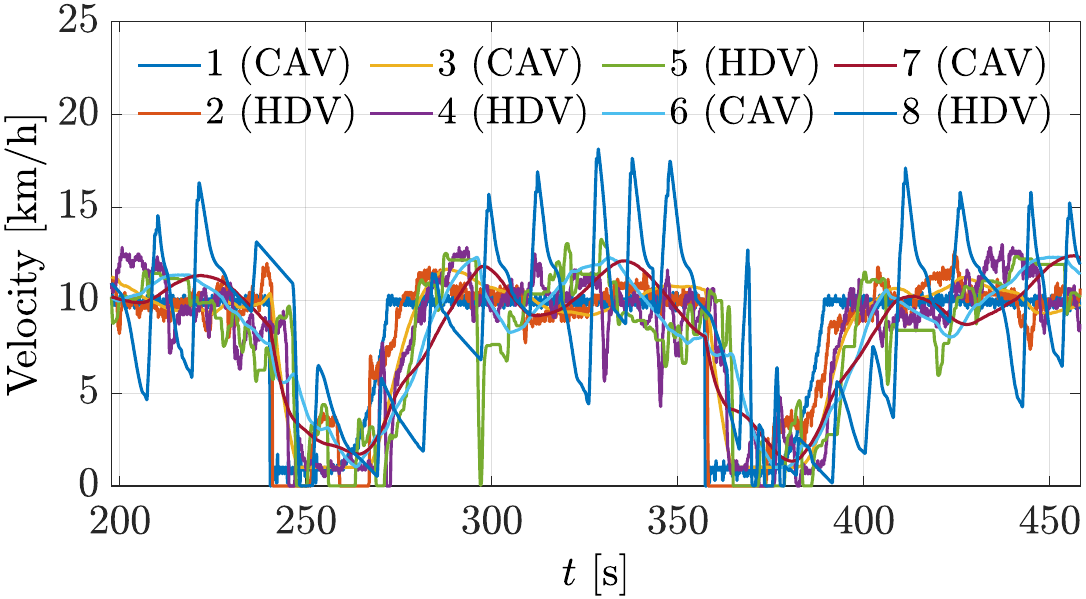}
		\label{fig-exp-velocity}}
	\hspace{4mm}
	\subfigure[Inter-vehicle distance profiles]
	{\includegraphics[width=0.45\textwidth]{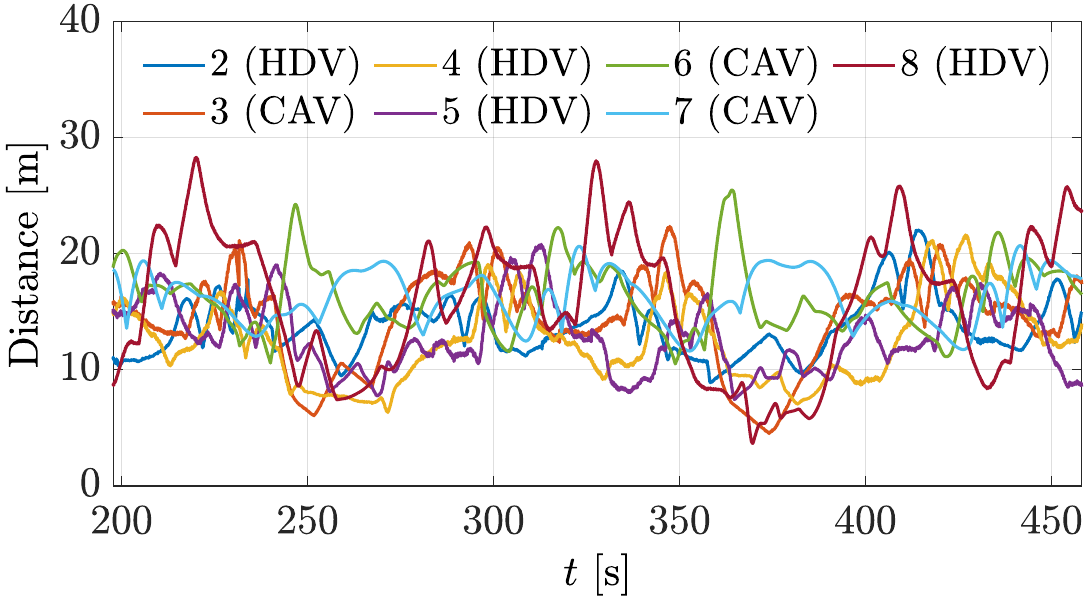}
		\label{fig-exp-distance}}
	\vspace{-1mm}
	\caption{Velocity and distance profiles of all vehicles in the experiments. Two braking disturbances are randomly applied to the head vehicle by the human operator through the MR HMD, which induces speed fluctuations throughout the entire platoon.
	}
	\label{fig-exp-result}
\end{figure*}

\subsection{Experimental Results}

The experimental results are presented in Fig.~\ref{fig-exp-result}, where velocity profiles are illustrated in Fig.~\ref{fig-exp-velocity}, and inter-vehicle distances are shown in Fig.~\ref{fig-exp-distance}.
The results demonstrate that the platoon composed of eight heterogeneous vehicles from multiple sources operates normally in the mixed testbed. 
Specifically, as shown in Fig.~\ref{fig-exp-velocity}, the braking perturbation randomly triggered by the human operator is accurately applied to the head vehicle, which induces speed fluctuations throughout the platoon. 
The CAVs distributed in the platoon effectively suppress the velocity disturbances caused by the preceding vehicles, preventing the propagation of oscillations along the platoon, which is consistent with the theoretical analysis~\cite{wang2021leading}.
Meanwhile, the authentic driving behaviors of human drivers are successfully captured. 
As shown in Fig.~\ref{fig-exp-result}, compared with CAVs, HDVs exhibit significantly larger speed fluctuations, as well as noticeable response delay and overshoot.

Furthermore, the braking perturbation constitutes a safety-critical corner case, which is challenging to realize in real-world traffic conditions. 
As shown in Fig.~\ref{fig-exp-distance}, the chain reaction triggered by sudden emergency braking reduces the inter-vehicle distance below $5 \mathrm{m}$, causing collisions in the experiment.
However, due to the interleaved distribution of physical and virtual vehicles within the platoon, no actual physical collision or hardware damage occur. 
This highlights how IMPACT’s “Physical-Virtual Action Interaction” capability facilitates safe VICS testing under corner cases involving that incorporates physical environments and entities.
In summary, these results validate  the effectiveness of the I-VIT platform in conducting   interactive VICS testing, further demonstrating the overall efficacy of the underlying IMPACT framework.

% Since human operators can directly interact with both physical and virtual entitiesin I-VIT, highly uncertain and unpredictable human behaviors are effectively incorporated into the testing to generate high-quality corner case. 
% Ultimately, VICS testing under safety-critical emergency braking corner cases is safely conducted.

\section{Conclusion}
\label{sec.5}

In this paper, we propose IMPACT, an interactive VICS testing framework built upon the mixedDT concept. 
% Beyond the L4 ``Optimizable'' level, we further propose the L5 ``Interactable'' level in the VICS-DT taxonomy, motivated by practical testing requirements.
Motivated by practical testing requirements, we introduce the L5 “Interactable” level in the VICS-DT taxonomy, extending beyond the existing L4 “Optimizable” level.
IMPACT is an L5-level framework that comprises physical, virtual, and mixed testing environments with the interactive environment, enabling human operators to directly interact with physical and virtual entities in VICS-DT systems.
This allows unpredictable and hard-to-model human behaviors to be directly incorporated into VICS testing, thereby naturally generating high-quality corner cases and serving as a valuable complement to AI-driven generation methods.
Furthermore, the mixedDT-enabled ``Physical-Virtual Action Interaction'' capability of IMPACT facilitates safe VICS testing under corner cases involving physical environments and entities. 
Finally, we implement the IMPACT framework through the I-VIT platform, and the conducted experiments validate the effectiveness of both the platform and the framework.

In the future, based on the IMPACT framework and the I-VIT platform, several research directions will be explored.
First, we intend to further investigate how MLLMs can empower VICS testing, especially in the generation of corner cases, and to conduct corresponding experiments on the I-VIT platform.
Second, we plan to further extend  the IMPACT framework, such as by refining the response and handling mechanisms for multiple concurrent interaction requests.
Finally, under the guidance of the IMPACT framework, we will conduct more extensive testing of VICS applications on I-VIT, including cooperative perception, cooperative control, and other VICS-based functionalities.

\section*{Acknowledgement}
This work is supported by National Natural Science Foundation of China, Science Fund for Creative Research Groups (52221005), and Tsinghua University-Didi Joint Research Center for Future Mobility.

%\section*{References}
	
\bibliography{mybibfile}

\end{document}